%% file: scifile.tex
\documentclass[12pt]{article}

\usepackage{times}

\input{sections/packages}

\newenvironment{sciabstract}{%
\begin{quote} \bf}
{\end{quote}}

\title{
Neural feels with neural fields:\\Visuo-tactile perception for in-hand manipulation
}

\newif\ifarxiv
\arxivtrue
\newif\ifwc

\author
{Sudharshan Suresh,$^{1, 2\ast}$ Haozhi Qi,$^{2, 3}$ 
Tingfan Wu,$^{2}$ Taosha Fan,$^{2}$ \\ 
Luis Pineda,$^{2}$ Mike Lambeta,$^{2}$ Jitendra Malik,$^{2, 3}$ Mrinal Kalakrishnan,$^{2}$ \\ 
Roberto Calandra,$^{4, 5}$ Michael Kaess,$^{1}$ Joseph Ortiz,$^{2}$ Mustafa Mukadam$^{2}$ \\
\\
\ifarxiv
    \normalsize{$^{1}$\href{https://www.ri.cmu.edu/}{CMU}},
    \normalsize{$^{2}$\href{https://ai.meta.com/research/}{FAIR}},
    \normalsize{$^{3}$\href{https://www.berkeley.edu/}{UC Berkeley}},
    \normalsize{$^{4}$\href{https://tu-dresden.de/?set_language=en}{TU Dresden}},
    \normalsize{$^{5}$\href{https://ceti.one/}{CeTI}}
\else
    \normalsize{$^{1}$CMU},
    \normalsize{$^{2}$FAIR},
    \normalsize{$^{3}$UC Berkeley},
    \normalsize{$^{4}$TU Dresden},
    \normalsize{$^{5}$CeTI}
\fi
\\
\\
\normalsize{$^\ast$To whom correspondence should be addressed; E-mail: \href{mailto:suddhus@gmail.com}{suddhus@gmail.com}.} 
}

\date{}

\ifarxiv
\else
    \usepackage{scicite}
\fi 

\ifarxiv 
    \usepackage[pagebackref=true]{hyperref}
\else
    \usepackage{hyperref}
\fi

\usepackage[capitalise]{cleveref}
\hypersetup{
    colorlinks,
    linkcolor={lcolor},
    citecolor={coolblack},
    urlcolor={lcolor}
}

\usepackage[utf8]{inputenc}
\usepackage[T1]{fontenc}
\usepackage{verbatim}

\begin{document} 

\input{sections/commands}

\baselineskip24pt

\maketitle 

\ifarxiv
{\setstretch{1.2}
\fi

\ifwc
\noindent Abstract: %
  \immediate\write18{texcount -1 -sum -merge -q sections/abstract.tex output.bbl > sections/abstract-words.sum }%
  (\input{sections/abstract-words.sum} words)%
 \\
Summary: %
  \immediate\write18{texcount -1 -sum -merge -q sections/summary.tex output.bbl > sections/summary-words.sum }%
  (\input{sections/summary-words.sum} words)%
 \\ 
Introduction: %
  \immediate\write18{texcount -1 -sum -merge -q sections/introduction.tex output.bbl > sections/introduction-words.sum }%
  (\input{sections/introduction-words.sum} words)%
 \\
Results: %
  \immediate\write18{texcount -1 -sum -merge -q sections/results.tex output.bbl > sections/results-words.sum }%
  (\input{sections/results-words.sum} words)%
 \\
Discussion: %
  \immediate\write18{texcount -1 -sum -merge -q sections/discussion.tex output.bbl > sections/discussion-words.sum }%
  (\input{sections/discussion-words.sum} words)%
 \\
Methods: %
  \immediate\write18{texcount -1 -sum -merge -q sections/methods.tex output.bbl > sections/methods-words.sum }%
  (\input{sections/methods-words.sum} words)%
 \\
Acknowledgements: %
  \immediate\write18{texcount -1 -sum -merge -q sections/acknowledgements.tex output.bbl > sections/acknowledgements-words.sum }%
  (\input{sections/acknowledgements-words.sum} words)%
 \\
Supplementary: %
  \immediate\write18{texcount -1 -sum -merge -q sections/supplementary.tex output.bbl > sections/supplementary-words.sum }%
  (\input{sections/supplementary-words.sum} words)%

\fi

\input{sections/abstract}
\ifwc
  \immediate\write18{texcount -1 -sum -merge -q sections/abstract.tex output.bbl > sections/abstract-words.sum }%
  (\input{sections/abstract-words.sum} words)%

\fi

\input{sections/summary}
\ifwc
  \immediate\write18{texcount -1 -sum -merge -q sections/summary.tex output.bbl > sections/summary-words.sum }%
  (\input{sections/summary-words.sum} words)%

\fi

\input{sections/introduction}
\ifwc
  \immediate\write18{texcount -1 -sum -merge -q sections/introduction.tex output.bbl > sections/introduction-words.sum }%
  (\input{sections/introduction-words.sum} words)%

\fi

\input{sections/results}
\ifwc
  \immediate\write18{texcount -1 -sum -merge -q sections/results.tex output.bbl > sections/results-words.sum }%
  (\input{sections/results-words.sum} words)%

\fi

\input{sections/discussion}
\ifwc
  \immediate\write18{texcount -1 -sum -merge -q sections/discussion.tex output.bbl > sections/discussion-words.sum }%
  (\input{sections/discussion-words.sum} words)%

\fi

\input{sections/methods}
\ifwc
  \immediate\write18{texcount -1 -sum -merge -q sections/methods.tex output.bbl > sections/methods-words.sum }%
  (\input{sections/methods-words.sum} words)%

\fi

\newpage
\bibliography{scibib}

\input{sections/acknowledgements}
\ifwc
  \immediate\write18{texcount -1 -sum -merge -q sections/acknowledgements.tex output.bbl > sections/acknowledgements-words.sum }%
  (\input{sections/acknowledgements-words.sum} words)%

\fi

\input{sections/supplementary}
\ifwc
  \immediate\write18{texcount -1 -sum -merge -q sections/supplementary.tex output.bbl > sections/supplementary-words.sum }%
  (\input{sections/supplementary-words.sum} words)%

\fi

\ifarxiv
}
\fi
\end{document}

%% file: sections/packages.tex
\usepackage{graphicx}
\graphicspath{{./images/}}
\usepackage{floatpag} %
\usepackage{gensymb} %

\usepackage{fullpage}
\usepackage{graphicx}
\usepackage{amsmath}

\usepackage[%
letterpaper,twoside,vscale=.8,hscale=.75,nomarginpar,hmarginratio=1:1
]{geometry}

\usepackage[T1]{fontenc}
\usepackage{epsfig} %
\usepackage{amsmath} %
\usepackage{amssymb}  %
\usepackage{color, soul} %
\usepackage{xcolor}
\usepackage{mathtools}
\usepackage{booktabs}
\usepackage{times}
\usepackage[labelfont=bf, font=footnotesize]{caption}%
\usepackage[labelfont=bf, font=footnotesize]{subcaption}%

\usepackage{lipsum}  
\usepackage{enumerate}
\usepackage{wrapfig}
\usepackage{balance}
\usepackage{algorithm}
\usepackage{setspace}
\usepackage{varwidth}%
\usepackage{algpseudocode}%
\usepackage{textcomp}
\usepackage{gensymb}
\usepackage{tabulary}
\usepackage{relsize}
\usepackage{graphbox} %

\usepackage{import}
\usepackage{tabularx}
\usepackage{array}
\usepackage{makecell}
\usepackage{stackengine}
\usepackage{multirow}
\usepackage{float}
\usepackage{xargs} %
\usepackage[export]{adjustbox}
\usepackage{inconsolata}
\usepackage{url}            %
\usepackage{enumitem}
\usepackage{blindtext}
\usepackage{regexpatch}
\usepackage{bm}
\usepackage{icomma}
\usepackage[title]{appendix}
\usepackage{tikz} %

\definecolor{brightpink}{rgb}{1.0, 0.0, 0.5}
\definecolor{ceruleanblue}{rgb}{0.16, 0.32, 0.75}
\definecolor{coolblack}{rgb}{0.0, 0.18, 0.39}
\definecolor{query}{HTML}{820D07}
\definecolor{meas}{HTML}{2AA63F}
\definecolor{factor}{HTML}{000287}

\definecolor{qcolor}{HTML}{FFA700}
\definecolor{icolor}{HTML}{075493}
\definecolor{pcolor}{HTML}{76D6FF}
\definecolor{fcolor}{HTML}{942193}
\definecolor{prcolor}{HTML}{929292}
\definecolor{lcolor}{RGB}{6,69,173}

\definecolor{contact_color}{HTML}{0B1BE2}
\definecolor{motion_color}{HTML}{FEA700}
\definecolor{shape_color}{HTML}{25A730}
\definecolor{shape_fill_color}{HTML}{D2E1D4}

\usepackage{xspace}

%% file: sections/commands.tex
\newcommand{\webpage}{https://suddhu.github.io/neural-feels}

\newcommand{\ours}{{N}eural{F}eels\xspace}%
\newcommand{\dataset}{{F}eel{S}ight\xspace}
\newcommand{\datasetreal}{\dataset~real\xspace}
\newcommand{\datasetsim}{\dataset~sim\xspace}

\newcommand{\simtoreal}{sim-to-real\xspace}
\newcommand{\etal}{et al.\xspace}
\newcommand{\App}{Appendix\xspace}
\newcommand{\Fig}{Figure\xspace}
\newcommand{\Sec}{Section\xspace}
\newcommand{\ie}{\textit{i}.\textit{e}., }
\newcommand{\eg}{\textit{e}.\textit{g}.\ }
\newcommand{\digitindex}{\texttt{t\_index}\xspace}
\newcommand{\digitmiddle}{\texttt{t\_middle}\xspace}
\newcommand{\digitring}{\texttt{t\_ring}\xspace}
\newcommand{\digitthumb}{\texttt{t\_thumb}\xspace}

%% file: sections/abstract.tex
\begin{sciabstract}
To achieve human-level dexterity, robots must infer spatial awareness from multimodal sensing to reason over contact interactions.
During in-hand manipulation of novel objects, such spatial awareness involves estimating the object's pose and shape. The status quo for in-hand perception primarily employs vision, and restricts to tracking a priori known objects. Moreover, visual occlusion of objects in-hand is imminent during manipulation, preventing current systems to push beyond tasks without occlusion. We combine vision and touch sensing on a multi-fingered hand to estimate an object's pose and shape during in-hand manipulation. Our method, \ours encodes object geometry by learning a neural field online and jointly tracks it by optimizing a pose graph problem. We study multimodal in-hand perception in simulation and the real-world, interacting with different objects via a proprioception-driven policy. Our experiments show final reconstruction F-scores of $\mathbf{81}$\% and average pose drifts of $\mathbf{4.7}\,\text{mm}$, further reduced to $\mathbf{2.3}\,\text{mm}$ with known CAD models. Additionally, we observe that under heavy visual occlusion we can achieve up to $\mathbf{94}$\% improvements in tracking compared to vision-only methods. Our results demonstrate that touch, at the very least, refines and, at the very best, disambiguates visual estimates during in-hand manipulation. We release our evaluation dataset of 70 experiments, \dataset, as a step towards benchmarking in this domain. Our neural representation driven by multimodal sensing can serve as a perception backbone towards advancing robot dexterity. Videos can be found on our project website: \url{\webpage}. 
\end{sciabstract}

%% file: sections/summary.tex
\section*{Summary}
\label{sec:summary}
\ifarxiv 
\vspace{-1mm}
\fi 
Neural perception with vision and touch yields robust tracking and reconstruction of novel objects for in-hand manipulation.

%% file: sections/introduction.tex
\begin{figure}[h]
    \thisfloatpagestyle{empty}
    \centering
    \includegraphics[width=\columnwidth,keepaspectratio]{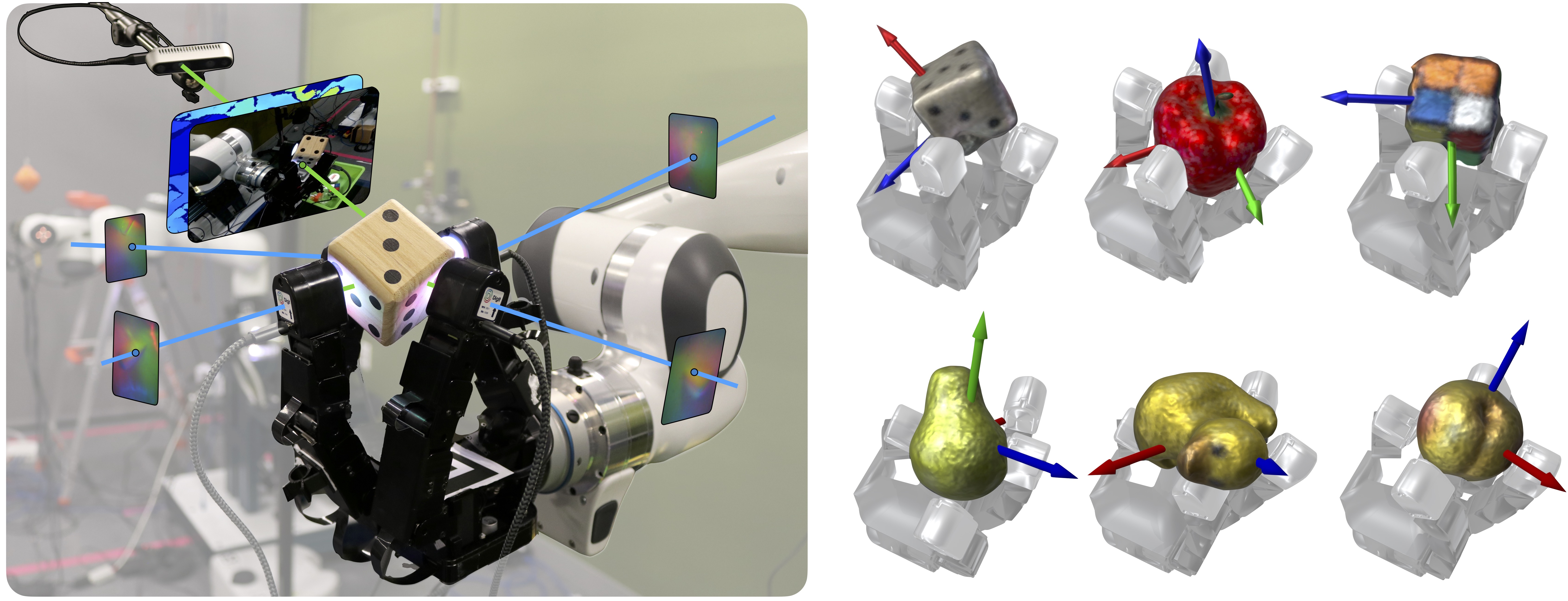}
    \caption{\textbf{Visuo-tactile perception with \ours.} Our method estimates pose and shape of novel objects \textbf{(right)} during in-hand manipulation, by learning neural field models online from a stream of vision, touch, and proprioception \textbf{(left)}.}
    \label{fig:cover}
\end{figure}

\ifarxiv
\vspace{-10mm}
\fi

\section{Introduction}
\label{sec:introduction}

To perceive deeply is to have sensed fully. Humans effortlessly combine their senses for everyday interactions---we can rummage through our pockets in search of our keys, and deftly insert them to unlock our front door. Currently, robots lack the cognition to replicate even a fraction of the mundane tasks we perform, a trend summarized by Moravec's Paradox~\cite{moravec1988mind}. For dexterity in unstructured environments, a robot must first understand its spatial relationship with respect to the manipuland.  Indeed, as robots move out of instrumented labs and factories to cohabit our spaces, there is a need for generalizable spatial AI~\cite{davison2018futuremapping}. 

Specific to in-hand dexterity, \textit{knowledge of object pose and geometry} is crucial to policy generalization~\cite{openai2018learning, openai2019solving, nvidia2022dextreme, qi2023general}.  As opposed to end-to-end supervision~\cite{yin2023rotating, guzey2023dexterity, chen2023visual}, these methods require a persistent 3D representation of the object. However, the status quo for in-hand perception is currently restricted to the narrow scope of tracking known objects with vision as the dominant modality~\cite{nvidia2022dextreme}. Further, it is common for practitioners to sidestep the perception problem entirely, retrofitting objects and environments with fiducials~\cite{openai2018learning, openai2019solving}. To further progress towards general dexterity, it is clear that one of the missing pieces is general, robust perception. 

With visual sensing, researchers tend to tolerate interaction rather than embrace it. This is at odds with contact-rich problems  where self-occlusions is imminent, like rotating~\cite{qi2023hand}, re-orienting~\cite{nvidia2022dextreme, chen2023visual}, and sliding~\cite{she2021cable, suresh2022midastouch}. Additionally, vision often fails in the real-world due to poor illumination, limited range, transparency, and specularity. Touch provides a direct window into these dynamic interactions, and human cognitive studies  have reinforced the complementarity with vision~\cite{helbig2007optimal}.

Hardware advances have led to affordable vision-based touch sensors~\cite{yuan2017gelsight, donlon2018gelslim, ward2018tactip, alspach2019soft, lambeta2020digit, padmanabha2020omnitact, wang2021gelsight} like the GelSight and DIGIT. Progress in touch simulation~\cite{wang2022tacto} enables practitioners to learn tactile observation models that transfer to real-world interactions~\cite{wang2021gelsight, sodhi2021patchgraph, suresh2022shapemap}. With a fingertip form-factor, their illuminated gel deforms on contact and the physical interaction is captured by an internal camera. When chained with robot kinematics, we obtain dense, situated contact that can be processed similar to natural camera images. 

Now given multimodal sensing, how best to represent the spatial information?  Coordinate-based learning, formalized as \textit{neural fields}~\cite{xie2022neural}, has found great success in visual computing. With neural fields, practitioners can create high-quality 3D assets offline given noisy visual data and pose annotation~\cite{mildenhall2021nerf, muller2022instant, li2023neuralangelo}. They are continuous representations with higher fidelity than their discrete counterparts like point clouds and meshes. While they are specialized towards batch optimization, lightweight SDF models~\cite{ortiz2022isdf, sucar2021imap, zhu2022nice, wen2023bundlesdf} have made online perception possible. 

Researchers have used this extensible architecture not just for continuous 3D quantities like signed distance fields (SDFs) and radiance~\cite{park2019deepsdf, mildenhall2021nerf, muller2022instant}, but also for pose estimation~\cite{yen2021inerf, wen2023bundlesdf}, planning~\cite{grote2023neural}, and latent physics~\cite{le2023differentiable}. Moreover, the ease of imparting generative priors~\cite{yu2021pixelnerf} and initializing with pre-trained models~\cite{park2019deepsdf} future-proofs them. While neural fields have emerged little by little in robot manipulation~\cite{zhong2022touching, kerr2022evo, wi2022virdo++, grote2023neural}, the optimization of multimodal data remains an open question.

The domain of our work---an intersection of simultaneous localization and mapping (SLAM) and manipulation---has been studied for over two decades. A first exemplar is from Moll and Erdmann~\cite{moll2004reconstructing}, who reconstruct the shape and motion of an object rolled between robot palms, later reproduced with specialized sensors~\cite{strub2014correcting, lepert2023hand}. Tactile SLAM has been thoroughly investigated for planar pushing due to its well-understood mechanics~\cite{yu2015shape, Suresh21tactile}. The combination of vision and touch has been explored for reconstructing fixed objects~\cite{wang20183d, smith20203d, suresh2022shapemap, chen2023sliding} and tracking known objects~\cite{yu2018realtime, lambert2019joint, sodhi2021learning}. Closest to our work is FingerSLAM~\cite{zhao2023fingerslam}, combining dense touch from a single finger with vision, however we consider the more challenging case of in-hand manipulation.

\textbf{\ours} presents an online solution to localize and reconstruct objects for in-hand manipulation with multimodal sensing. We unify vision, touch, and proprioception into a neural representation and demonstrate SLAM for apriori unknown objects, and robust tracking of known objects. In our experiments, we present our robot with a novel object, and it infers and tracks its geometry through just interaction. We use a dexterous hand~\cite{allegro2023} sensorized with commercial vision-based touch sensors~\cite{lambeta2020digit} and a fixed RGB-D camera (\Fig \ref{fig:cover}). With a proprioception-driven policy~\cite{qi2023hand} we explore the object's extents through in-hand rotation.

Through our experiments we study the role that vision and touch play in interactive perception, the effects of occlusion, and visual sensing noise. To evaluate our work, we collect a dataset of $70$ in-hand rotation trials in both the real-world and simulation, with ground-truth object meshes and tracking. Our results on novel objects show average reconstruction F-scores of $81\%$ with pose drifts of just $4.7\,\text{mm}$, further reduced to $2.3\,\text{mm}$ with known CAD models. Under heavy occlusion, we demonstrate up to $94$\% improvements in pose tracking compared to vision-only methods. Our combination of rich sensing and spatial AI requires minimal hardware compared to complex sensing cages, and is easier to interpret than end-to-end perception methods. The output of the neural SLAM pipeline---pose and geometry---can drive further research in general dexterity, broadening the capabilities of home robots.

%% file: sections/results.tex
\begin{figure}[t!]
    \thisfloatpagestyle{empty}
    \centering
    \href{https://suddhu.github.io/neural-feels/video/pipeline.m4v}{\includegraphics[width=\columnwidth,keepaspectratio]{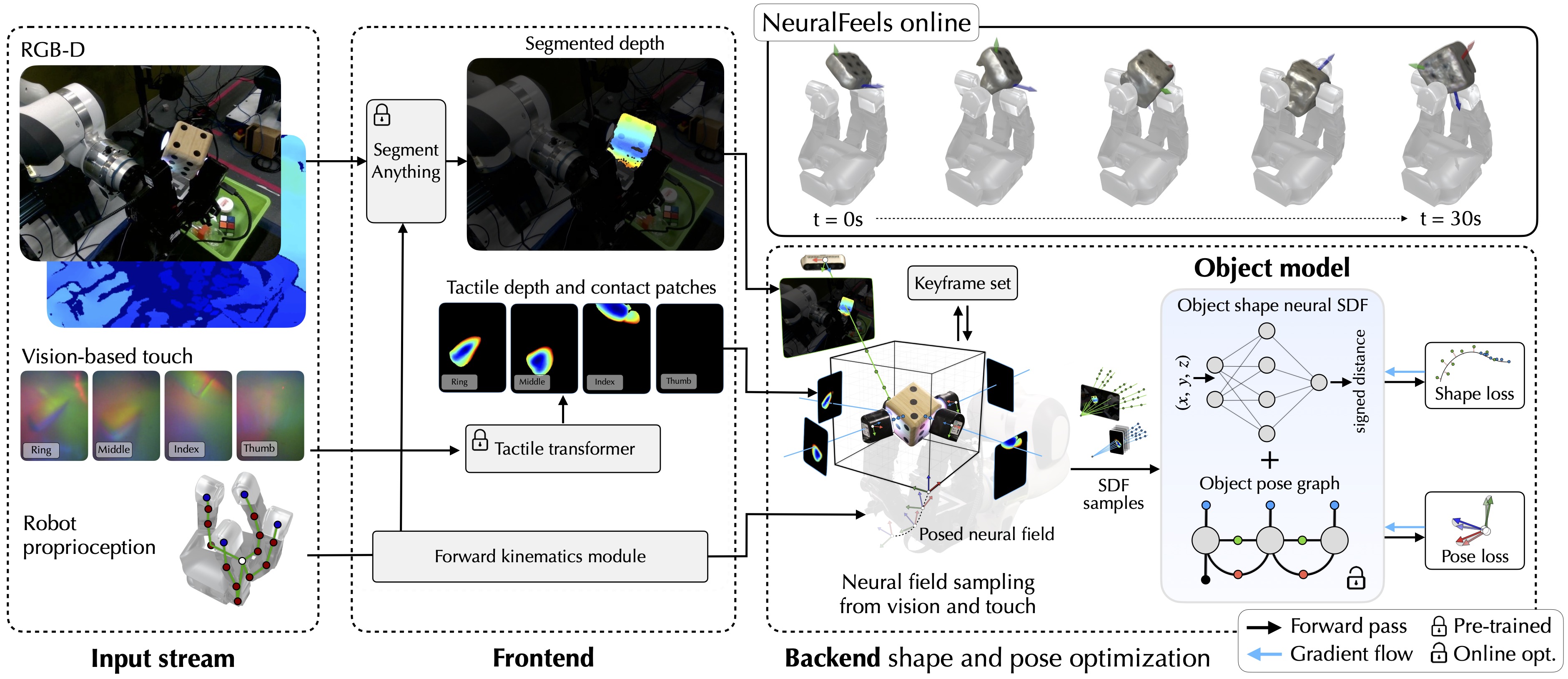}}
    \caption{\textbf{A visuo-tactile perception stack amidst interaction.} An online representation of object shape and pose is built from vision, touch, and proprioception during in-hand manipulation. Raw sensor data is first fed into the \textit{frontend}, which extracts visuo-tactile depth with our pre-trained models. Following this, the \textit{backend} samples from the depth to train a neural signed distance field (SDF), while the pose graph tracks the posed neural field.}
    \label{fig:pipeline}
    \ifarxiv 
    \vspace{-1em}
    \fi 
\end{figure}

\section{Results}
\label{sec:results}

Our multi-fingered robot hand is presented with a novel object, placed randomly between its fingertips. It rotates the object in-hand, through a proprioception-driven policy~\cite{qi2023hand}, which gives rise to a stream of visual and tactile signals. We combine the visual, tactile, and proprioceptive sensing into our online neural field, for a persistent, evolving 3D representation of the unknown object. The full pipeline of our \ours perception stack is illustrated in \Fig \ref{fig:pipeline}. We also summarize our experiments and findings in \href{\webpage}{our webpage}.

We evaluate \ours over simulated and real-world interactions, totaling up to $70$ experiments over different object classes. Details of the dataset can be found in \Sec \ref{ssec:feelsight}. First, we demonstrate SLAM results for novel objects, and highlight some qualitative examples. Next, we demonstrate pose-tracking when we have a priori shape of the manipuland. Finally, we analyze the role touch plays in improving perception under occlusion and visual sensing noise. 

\subsection{Metrics and baseline}
\label{ssec:metrics_and_baseline}

\textbf{Pose and reconstruction metrics.} We use the symmetric average Euclidean distance metric (ADD-S) to evaluate the pose tracking error over time~\cite{tremblay2023diff}. The ADD metric is commonly used in manipulation~\cite{xiang2017posecnn, bauza2019tactile, tremblay2018deep, tremblay2023diff} as a geometrically-interpretable distance metric for pose error. It is computed by sub-sampling the ground-truth object mesh and averaging the Euclidean distance between the point-set in the estimated and ground-truth object pose frames. Rather than pairwise distance, ADD-S considers the closest point distance, which disambiguates symmetric objects. 

For reconstruction, we compare how \textit{accurate} (precision) and \textit{complete} (recall) the neural SDF is in comparison to the ground-truth mesh. The F-score, an established metric in the multi-view reconstruction community~\cite{knapitsch2017tanks, tatarchenko2019single}, combines these two criteria into an interpretable ${\small [0\!-\!1]}$ value. To compute this, we first sub-sample the ground-truth and reconstructed meshes, and transform both to the common object-centric reference frame. Given a distance threshold, in our case $\tau\!=\!5\,\text{mm}$, \textit{precision} measures the percentage of reconstructed points within $\tau$ distance from the ground-truth points. Conversely, \textit{recall} measures the percentage of ground-truth points within $\tau$ distance from the reconstructed points. The harmonic mean of these two quantities give us the F-score, which captures both surface reconstruction accuracy and shape completion. Broadly, a higher F-score with tighter $\tau$ bounds implies better object reconstructions. For brevity, we refer to ADD-S and F-score as the \textit{pose metric} and \textit{shape metric} respectively. 

\textbf{Ground-truth shape and pose.} We evaluate these metrics against the ground-truth estimates of object shape and pose. For each object, the ground-truth shape is obtained from offline scans (\Fig \ref{fig:ground_truth}). Ground-truth object pose is straightforward in simulation experiments, directly exposed by IsaacGym~\cite{makoviychuk2021isaac}. In the real-world, we estimate a pseudo ground-truth, via multi-camera pose tracking of the experiment. Instrumented solutions, such as 3D motion capture, are infeasible as it both visually and physically interferes with the experiments. We opt to install two additional cameras (\Sec \ref{ssec:feelsight}) and run \ours in pose tracking mode with the ground-truth object shape. This represents the \textit{best tracking estimates given known shape and occlusion-free vision}. For further details, refer to \Sec \ref{ssec:gt_shape_pose}. 

\subsection{Neural SLAM: object pose and shape estimation}
\label{ssec:neural_slam}

In this section, we evaluate \ours' ability for embodied spatial reasoning from scratch. We present the robot with a novel object, and the robot is tasked with building an object model on-the-fly. This is typical where robots continually learn from interaction, such as when deployed in unstructured household environments. We make no assumptions about the object geometry, which is built from scratch, or manipulation actions, which are decided at deployment. We process visuo-tactile data sequentially with no access to future information or category-level priors. This formulation aligns with other dexterous manipulation work~\cite{nvidia2022dextreme, qi2023hand, qi2023general, chen2023visual}, and is less restrictive than that of FingerSLAM~\cite{zhao2023fingerslam}, where the object is always in contact with a single tactile sensor and the camera is unobstructed. 

\begin{figure}[t]
    \thisfloatpagestyle{empty}
    \centering
    \includegraphics[width=\columnwidth,keepaspectratio]{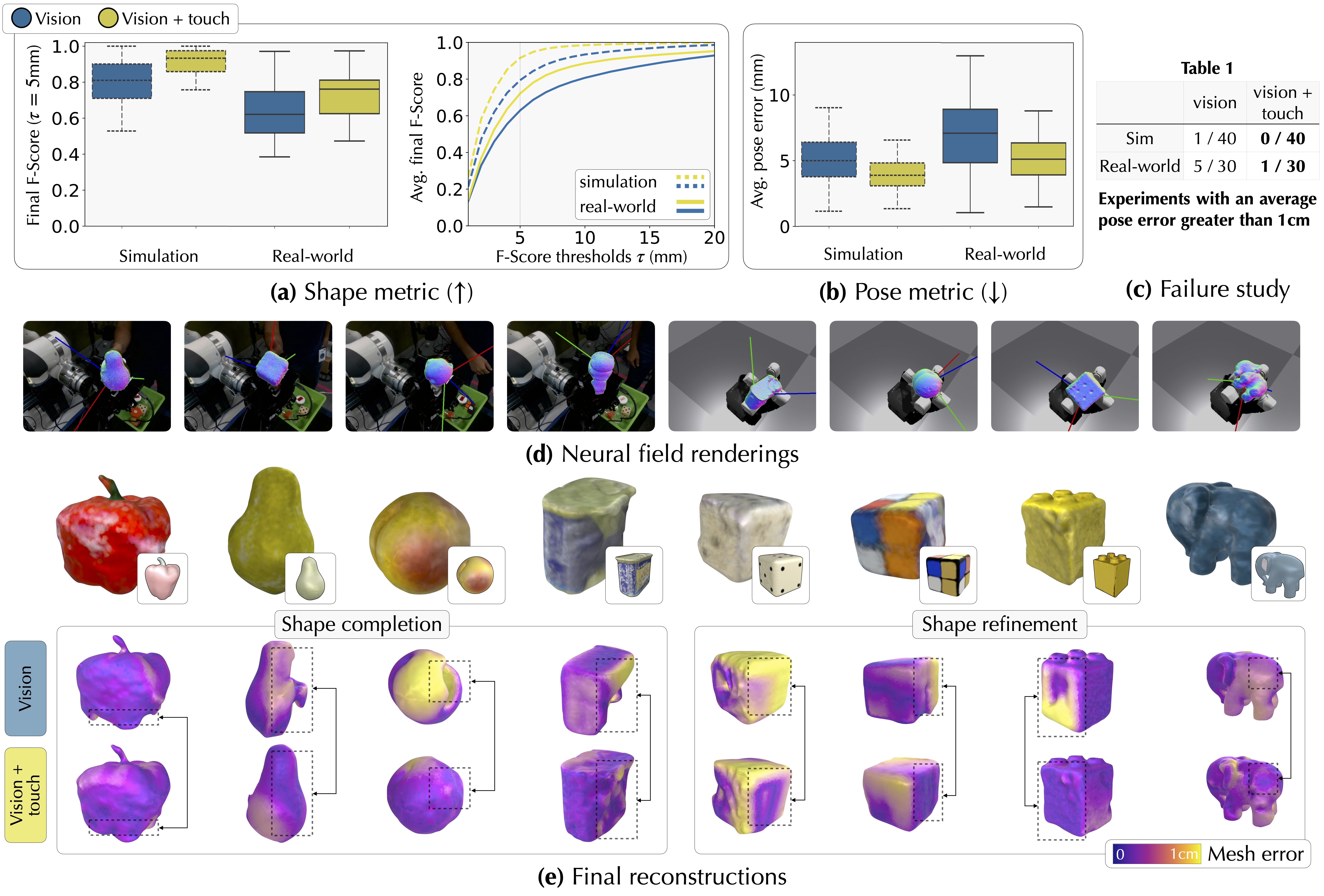}
    \caption{\textbf{Summary of SLAM experiments.} \textbf{(a, b)} We present aggregated statistics for SLAM over a combined 70 experiments (40 in simulation and 30 in the real-world), with each trial run over 5 different seeds. We compare across simulation and real-world to show low pose drift and high reconstruction accuracy. \textbf{(c)} Table 1 illustrates the number of trials that our method fails to track (and reconstruct) the object. \textbf{(d)} Representative examples of the final object pose and neural field renderings from the experiments. \textbf{(e)} The final 3D objects generated by marching cubes on our neural field. Here, we highlight the role tactile plays in both shape completion and shape refinement.}
    \label{fig:aggregate_slam}
    \ifarxiv 
    \vspace{-1em}
    \fi 
\end{figure}

We evaluate over a combined $70$ experiments in simulation and real-world across of $14$ different objects. The objects are placed in-hand, after which the policy collects $30$ seconds of vision, touch, and proprioception data. As each run is non-deterministic, we average our results across $5$ different seeds, resulting in a total of $350$ trials. The first frame of each sequence only presents limited visual knowledge: a single side of \textit{Rubik's cube} or \textit{large dice}; the underside of the \textit{rubber duck}. Through the course of any $30$ second sequence, in-hand rotation exposes previously unseen geometries to vision and touch fills in the rest of the occluded surfaces. In \Fig \ref{fig:aggregate_slam}, we present the main set of results, where we compare the multimodal fusion schemes against ground-truth. 

\textbf{Object reconstructions.} \Fig \ref{fig:aggregate_slam} (a) shows the final shape metric at the end of each sequence for a fixed threshold $\tau$. Here we pick $\tau\!=\!5\,\text{mm}$ for this evaluation, around $3$\% of the maximum diagonal length of the objects. Greater the value of the shape metric, the closer the surface reconstructions are to ground-truth. We observe large gains when incorporating touch, with surface reconstructions on average $15.3$\% better in simulation and $14.6$\% better in the real-world. Our final reconstructions, as seen in \Fig \ref{fig:aggregate_slam} (e), have a median error of $2.1\text{mm}$ in simulation and $3.9\text{mm}$ in the real-world. Additionally, the second plot compares the final shape metrics against a range of $\tau$ thresholds. Here we observe that multimodal fusion leads to consistently better shape metrics across all $\tau$ values in simulation and the real-world. 

\textbf{Object pose drift.} As SLAM is the exemplar of a chicken and egg problem, there is a strong correlation between a low shape metric and high pose metric. Empirically, we observe larger pose drift in the initial few seconds due to incomplete geometry, which levels off with further exploration. For fair comparisons we initialize the object's canonical pose to the ground-truth, but this is not necessary otherwise. With this initialization, we ignore the pose metric over the first five seconds, as it is ill-defined.  

\Fig \ref{fig:aggregate_slam} (b) plots the drift of the object's estimated pose with respect to the ground-truth, lower being more accurate. We observe better tracking with respect to the vision-only baseline, with improvements of $21.3$\% in simulation and $26.6$\% in the real-world. Table 1 in \Fig \ref{fig:aggregate_slam} (c) reports the number of failures in vision-only tracking compared to \ours. Here, a failed experiment is defined as when the average pose drift exceeds an empirical threshold of $10\,\text{mm}$. 

\textbf{Qualitative results.} \Fig \ref{fig:aggregate_slam} (d) visualizes the rendered normals of the posed neural field at the end of each experiment, with the 3D coordinate axes superimposed. The final 3D reconstructions, generated via marching cubes, are shown in \Fig \ref{fig:aggregate_slam} (e) alongside the ground-truth meshes. Below that, we highlight the gains with visuo-tactile integration, with examples of shape completion and refinements. 

\begin{figure}[H]
    \thisfloatpagestyle{empty}
    \centering
    \includegraphics[height=0.95\textheight,keepaspectratio]{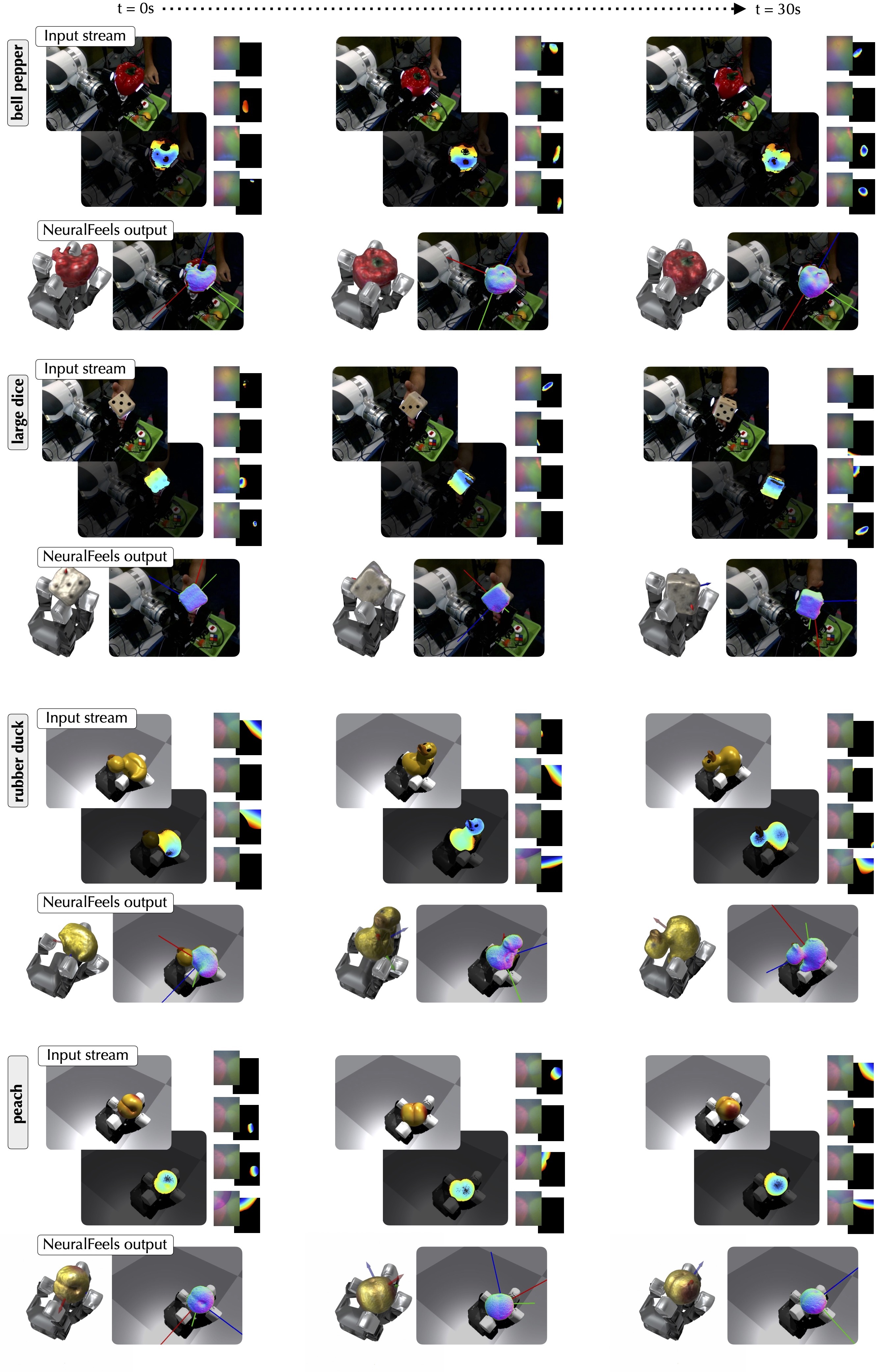}
    \caption{\textbf{Representative SLAM results.} In both real-world and simulation, we build an evolving neural SDF that integrates vision and touch while simultaneously tracking the object. We illustrate the input stream of RGB-D and tactile images, paired with the posed reconstruction at that timestep.}
    \label{fig:slam_real}
\end{figure}

In \Fig \ref{fig:slam_real} we show the incremental pose tracking and reconstructions of objects across different time slices of a few representative experiments. We present two results from the real-world, \textit{bell pepper} and \textit{large dice}, and two from simulation, \textit{rubber duck} and \textit{peach}. At each timestep, we highlight the input stream, frontend depth and output object model. The 3D visualizations are generated by marching-cubes, in addition to the rendered normals of the neural field projected onto the visual image. In each case, we partially reconstruct the object at the initial frame, and build the surfaces out progressively over time. 

\subsection{Neural tracking: object pose estimation given shape}
\label{ssec:neural_pose}

As a special case of \ours, we demonstrate superior multimodal pose tracking when provided the CAD models of the objects at runtime. Tracking known geometries is an active area of research in visual SLAM~\cite{nvidia2022dextreme, labbe2022megapose}, with some work that incorporates touch as well~\cite{yu2018realtime, lambert2019joint, sodhi2021learning, suresh2022midastouch, bauza2023simple}. This is applicable in environments like warehouses and manufacturing lines, where robots have intimate knowledge of the manipulands~\cite{bauza2023simple}. It is further useful in household scenarios, where the robot has already generated an object model through interaction.

In implementation, the object's SDF is pre-computed from a given CAD model. During runtime, we freeze the weights of the neural field, and only perform visuo-tactile tracking with the frontend estimates. Similar to the SLAM experiments, we run each of the $70$ experiments over $5$ seeds, and report the pose metrics with respect to ground-truth.

\begin{figure}[t]
    \thisfloatpagestyle{empty}
    \centering
    \includegraphics[width=\columnwidth,keepaspectratio]{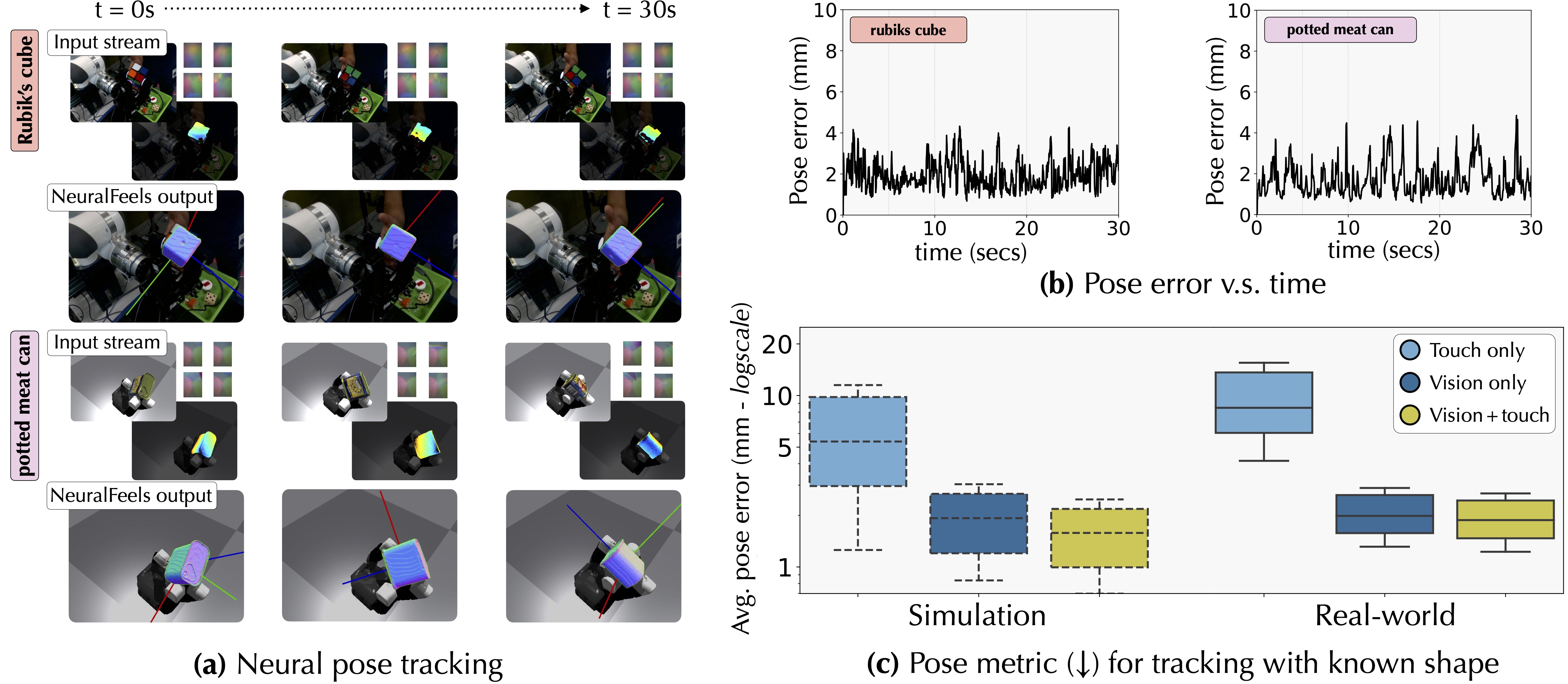}
    \caption{\textbf{Neural pose tracking of known objects.} \textbf{(a)} With known ground-truth shape, we can robustly track objects such as the Rubik's cube and potted meat can. \textbf{(b)}  We observe reliable tracking performance, with average pose errors of $2\,\text{mm}$ through the sequence. \textbf{(c)} With a known object model and good visibility,  touch plays the role of pose refinement.}
    \label{fig:representative_pose}
    \ifarxiv 
    \vspace{-5mm}
    \fi 
\end{figure}

\textbf{Results from pose tracking.} \Fig \ref{fig:representative_pose} (a) shows some qualitative examples of tracking the pose of the \textit{Rubik's cube} and \textit{potted meat can} with vision and touch. For the given examples, the pose metric over the sequences are plotted in \Fig \ref{fig:representative_pose} (b). We observe low, bounded pose error even with imprecise visual segmentation and sparse touch signals. In \Fig \ref{fig:representative_pose} (c) we observe the role touch plays in reducing the average pose error over all experiments to the range of $2.3\,\text{mm}$. Given the CAD model, we observe that incorporating touch can refine our pose estimates, with a decrease in average pose error by $22.29$\% in simulation and $3.9$\% in the real-world. As addresed in \Sec \ref{sec:discussion}, the less-pronounced contacts in the real-world can explain this disparity. In the following section, we highlight greater improvements with respect to the baseline when visual sensing is suboptimal.

\begin{figure}[t]
    \thisfloatpagestyle{empty}
    \centering
    \includegraphics[width=\columnwidth,keepaspectratio]{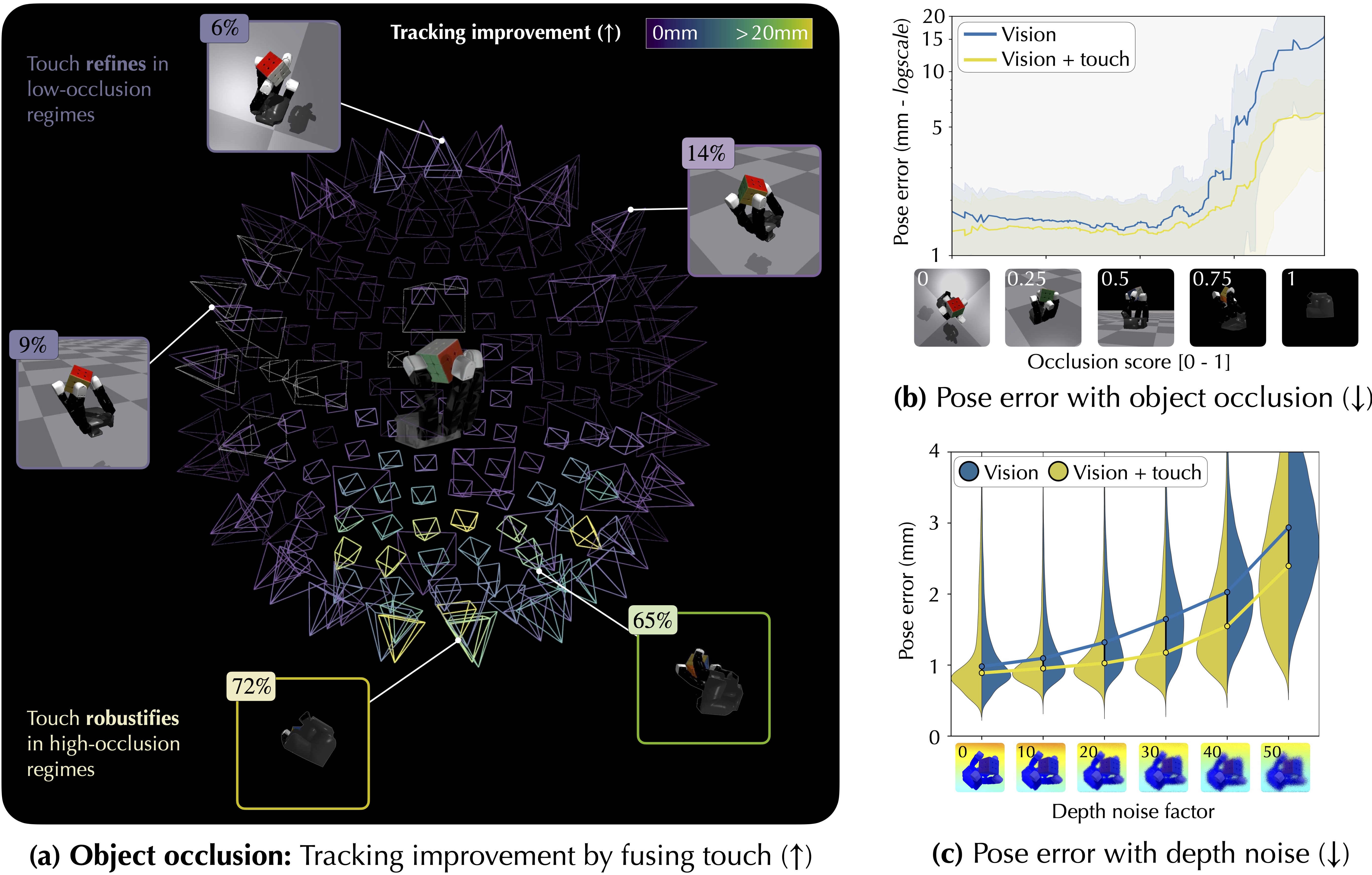}
    \caption{\textbf{Ablations on occlusions and sensing noise.} \textbf{(a)} With occluded viewpoints, visuo-tactile fusion helps improve tracking performance with an unobstructed local perspective. We quantify these gains across a sphere of camera viewpoint to show improvements, particularly in occlusion-heavy points-of-view. \textbf{(b)} We observe that touch plays a larger role when vision is heavily occluded, and a refinement role when we there is negligible occlusion. \textbf{(c)} With larger noise in visual depth, tactile help curb large pose tracking errors. }
    \label{fig:noise_ablation}
\end{figure}
\subsection{Perceiving under duress: occlusion and visual depth noise}
\label{ssec:occlusion}
In this section, we explore the broader benefits of fusing touch and vision through ablations on visual sensing properties. The previous results were achieved through the iterative co-design of perception and hardware, such that we have favorable camera positioning and precise stereo depth tuning. Indeed, this attention to detail is necessary for practitioners~\cite{nvidia2022dextreme, chen2023visual}, \textit{but can we also harness touch to improve over sub-optimal visual data?} We consider two such scenarios in simulation, where we can freely control these parameters, and evaluate on the pose tracking problem from the previous section.

\textbf{The effects of camera-robot occlusion.} In an embodied problem, third-person and egocentric cameras are both susceptible to occlusion from robot motion and environment changes. For example, if we were to retrieve a cup off the top shelf in the kitchen, we rely primarily on tactile signals to complete the task. For the perception system, this translates to the object of interest disappearing from the field of view, while local touch sensing is still unaffected. To emulate this we consider tracking the pose of a known \textit{Rubik's cube}. We simulate $200$ different cameras in a sphere of radius $0.5\,\text{m}$, each facing towards the robot. As shown in \Fig \ref{fig:noise_ablation} (a), each camera captures a unique vantage point of the same in-hand sequence, with varying levels of robot-object occlusion. This serves as proxy for occlusion faced by an egocentric or fixed camera when either the hand or environment occludes the object. 

To simplify the experiment, we assume the upper-bound performance of the vison-only frontend by providing ground-truth object segmentation masks. We characterize the visibility in terms of an \textit{occlusion score} by calculating the average segmentation mask area for each viewpoint, and normalizing them to $\left[0\!-\!1\right]$. For example, scores closer to $0$ correspond to viewpoints beneath the hand (most occluded), while those closer to $1$ correspond to cameras placed atop (least occluded). We run pose tracking experiments for each of the $200$ cameras in two modes: vision-only and visuo-tactile and compare between them.

In \Fig \ref{fig:noise_ablation} (a) we colormap each camera view based on the pose tracking improvements from incorporating touch. On average the improvement across all cameras is $21.2$\%, and it peaks at $94.1$\% at heavily occluded views. We inset frames from a few representative viewpoints and their corresponding relative improvement with visuo-tactile fusion. In \Fig \ref{fig:noise_ablation} (b) the pose error for each modality is further plotted versus the $\left[0\!-\!1\right]$ occlusion score. This corroborates the idea that touch refines perception in low-occlusion regimes and robustifies it in high-occlusion regimes. 

\textbf{The effects of noisy visual depth.} Depth from commodity RGB-D sensors are degraded as a function of camera-robot distance, environment lighting, and object specularity. Even in ideal scenarios, the RealSense depth algorithm has $35$ hyperparameters~\cite{keselman2023optimizing} that considerably affect the frontend input to \ours.  To simulate this, we corrupt the depth maps progressively with a realistic RGB-D noise, and observe the tracking performance for a known geometry. 

As implemented by Handa~\etal~\cite{handa2014benchmark}, we simulate common sources of depth-map errors as a sequence of pixel shuffling, quantization, and high frequency noise. The depth noise factor $D$ determines the magnitude of these operations, with the depth-maps visualized in \Fig \ref{fig:noise_ablation}~(c). While all prior simulation experiments have been collected with $D\!=\!5$, here we vary the magnitude from $0\!-\!50$ in intervals of $10$. At each noise level, we run pose tracking across the $5$ Rubik's cube experiments with $5$ unique seeds, resulting in a total of $150$ experiments. In \Fig \ref{fig:noise_ablation} (c) we plot error against the noise factor $D$, showing an expected upward trend in error with noise. However, we see markedly better tracking when fusing touch, especially in high-noise regimes. 

%% file: sections/discussion.tex
\section{Discussion}
\label{sec:discussion}

\noindent \textbf{\ours achieves robust object-centric SLAM through interaction.} To the best of our knowledge, \ours is the first demonstration of full-SLAM for multimodal, multifinger manipulation. We are inspired by computer vision systems that achieve high-fidelity neural reconstructions without pose annotation~\cite{sucar2021imap, zhu2022nice, wen2023bundlesdf} through online learning. They highlight the benefit of co-designed pose tracking and reconstruction, which has also shown promise in manipulation systems~\cite{Suresh21tactile, zhao2023fingerslam}. More broadly, our stack relies on recent progress in somewhat disparate fields: SLAM, neural rendering, tactile sensing, and reinforcement-learning. 

As shown in the \Fig \ref{fig:aggregate_slam} (a), we achieve average reconstruction F-scores of $81$\% across simulation and real-world experiments on novel objects. Simultaneously, we stably track these objects amidst interaction with minimal drift, an average of  $4.7\,\text{mm}$. While the vision-only baseline may suffice for some scenarios, the results validate the utility of rich, multimodal sensing for interactive tasks. This corroborates years of research in interactive perception from touch and vision~\cite{smith2021active, suresh2022shapemap, bauza2023simple}, now applied on dexterous manipulation platforms. 

\ifarxiv 
\vspace{3mm}
\fi 

\noindent \textbf{Touch and proprioception ground embodied perception.} Interactive perception is far from ideal, an embodiment can more often than not get in the way of sensing. As seen in \Fig \ref{fig:slam_real}, in-hand manipulation suffers from challenges such as frequent occlusions, limited field-of-view, noisy segmentation, and rapid object motion. To tackle this, proprioception helps focus the perception problem: we can accurately singulate the object of interest through embodied prompting (\Sec \ref{sssec:segmented_vision}). When combined with touch, we robustify our visual estimates by giving us a window into local interactions. These are evident in simulated / real SLAM and pose tracking experiments, where multimodal fusion leads to improvements of $15.3$\% / $14.6$\% in reconstruction and $21.3$\% / $26.6$\% in pose tracking. 

Qualitatively, we see touch performs two key functions: \textbf{(i)} disambiguating noisy frontend estimates and \textbf{(ii)} providing context in the presence of occlusion. The former alleviates the effect of noisy visual segmentation and depth with co-located local information for mapping and localization. The latter provides important context hidden from visual sensing, like the occluded face of the \textit{large dice} or back of the \textit{rubber duck}. The final reconstructions in \Fig \ref{fig:aggregate_slam} (e) support these findings, with improved shape completion and refinement. This is important in the few-shot interactions of everyday life, where the richer sensing can create better object models.

The largest gains from incorporating touch are in heavy-occlusion regimes (\Fig \ref{fig:noise_ablation} (a)), where we can observe up to $94.1$\% improvements at certain camera viewpoints. To our knowledge, this co-design of perception and hardware has not been explored by practitioners before. This doesn't just demonstrate the complementary nature of the modalities, but further, the ideal configurations for occlusion-free manipulation. Finally, our results in tactile-only tracking (\Fig \ref{fig:representative_pose} (c)) support the analysis of Smith \etal~\cite{smith20203d} that learning exclusively from touch leads to poor performance as it lacks any global context. 

\ifarxiv 
\vspace{3mm}
\fi 

\noindent \textbf{Modularity marries pre-training with online learning.} As opposed to an end-to-end perception, \ours is fully interpretable due to its modular construction. This allows us to combine foundational models trained on large-scale image and tactile data (\textit{frontend}), with SLAM as online learning (\textit{backend}). Furthermore, our backend is a combination of state-of-the-art neural models~\cite{muller2022instant} with classical least-square optimization~\cite{pineda2022theseus} that have found success in SLAM~\cite{cadena2016past}. Chaining these systems together, we can achieve first-of-its-kind multimodal SLAM results without explicit training in the domain. This is crucial given the dearth of training data for in-hand tasks, and robot manipulation in general. 

This modular design has benefits for future generalization of our system: \textbf{(i)} Other models of tactile sensors~\cite{yuan2017gelsight, wang2021gelsight, alspach2019soft} can be easily integrated as long as they can be accurately simulated; \textbf{(ii)} alternate scene representations~\cite{barron2021mip, kerbl20233d} can supplant our neural field model, as required; \textbf{(iii)} additional state knowledge can be seamlessly integrated as factor graph cost functions, \eg tactile odometry~\cite{zhao2023fingerslam} and force-constraints~\cite{Suresh21tactile}; \textbf{(iv)} any combination of tactile and visual sensors can be fused into our multimodal framework with appropriate calibration and kinematics. 

\ifarxiv 
\vspace{3mm}
\fi 

\noindent \textbf{Application towards perception-driven planning.}  \ours is relevant to manipulation researchers and practitioners who require spatial perception with a single camera and affordable tactile sensing. It can be extended to not just in-hand rotation, but many other object-centric manipulation tasks like in-hand reorientation~\cite{chen2023visual}, pick-and-place~\cite{bauza2023simple}, insertion~\cite{lepert2023hand}, nonprehensile sliding~\cite{kerr2022learning}, and planar pushing~\cite{Suresh21tactile}. In the future, we hope to generalize to these different tasks and varied robot morphologies. While not explored in this work, the direct benefit of an online SDF is the ability to seamlessly plan for dexterous interactions. Recent works demonstrate the benefit of apriori-known object point-clouds~\cite{qi2023general} and SDFs~\cite{driess2022learning} for goal-conditioned planning, and running our perception stack in-the-loop is the next natural step. 

\ifarxiv 
\vspace{3mm}
\fi 

\noindent \textbf{System limitations.} \ours shows the potential of a multimodal system for manipulation that leverages pre-training and online learning for high accuracy spatial understanding. We present some of the limitations and promising directions for future work: 
\begin{itemize}
    \setlength{\itemsep}{-5pt} %
    \item \textbf{Generic 3D priors for object reconstruction.} For each experiment with a novel object, our method learns a 3D geometry \textit{from scratch} to best explain the visuo-tactile sensor stream. The pose tracker has a higher chance of failure in the initial few seconds, when the neural SDF is a poor-approximation of the full object due to limited sensor coverage. We further note that our rotation policy might not completely explore the object in the real-world, resulting in a lower average final F-Score of $81$\%. Out-of-scope in our work, but of great interest in the visual learning community~\cite{park2019deepsdf, wu2023multiview, hong2023lrm}, is leveraging pre-trained models for an initial object prior. Given an initial occluded view, careful integration of these large reconstruction models trained via category~\cite{park2019deepsdf} or multi-view supervision~\cite{wu2023multiview, hong2023lrm} may yield an initial-guess SDF that we refine over time with vision and touch. In manipulation, Wang \etal~\cite{wang20183d} have seen promising results in using shape priors for visuo-tactile reconstruction of fixed objects. 
    \item  \textbf{\expandafter\MakeUppercase \simtoreal adaptation.} Our findings indicate that while multimodal fusion performs well both in simulation and the real-world, its benefits are less pronounced in real-world deployment. This is a common problem in \simtoreal applications, and we qualitatively identify several domain gaps that explain this: \textbf{(i)} the DIGIT elastomer is less sensitive in real-world deployment, leading to sparser contact predictions; \textbf{(ii)} our RL policy is less stable in the real-world (sometimes requiring human intervention) and causes rapid jumps in object motion; \textbf{(iii)} noise in proprioception is only indirectly modelled as uncertainty terms in estimation. To tackle these, we must leverage work in \simtoreal generalization for tactile simulation~\cite{higuera2023learning} and reinforcement-learning~\cite{qi2023general}. 
    \item  \textbf{System design considerations.} We identify viable engineering improvements that can be made towards a general-purpose system. We are currently restricted to a fixed-camera setup, with an online hand-eye calibration or egocentric vision, this can be relaxed. Depth uncertainty~\cite{dexheimer2023learning} is valuable information for our neural model to handle visually-adversarial objects like glass and metal. To achieve true real-time frequencies, efficiency gains can be made in the pose optimizer and frontend estimation. Finally, we can increase robustness by using the color information for feature-based tracking of objects~\cite{detone2018superpoint}. 
    \item \textbf{Perceiving latent state.} We consider geometry as just the starting point for neural models: interaction reveals latent properties like texture~\cite{kerr2022learning}, friction~\cite{le2023differentiable}, and object dynamics~\cite{sundaralingam2021hand}. Neural fields can embed these latents as auxiliary optimization terms so as to  benefit tasks that go beyond just geometry and pose. Applications can range from learning to manipulate inertially-significant objects (\eg a hammer), to identifying a grasp point from local texture (\eg a saucepan handle). 
    \end{itemize}

%% file: sections/methods.tex
\section{Materials and methods}
\label{sec:materials}

\ours ingests multimodal information to build a persistent 3D object representation. Similar to classical SLAM frameworks, it first has a \textit{frontend}, responsible for abstracting the vision (RGB-D) and touch (RGB) input stream into a format suitable for estimation (segmented depth). Thereafter, the \textit{backend} fuses this data into an optimization structure that infers the \textit{object model}: an evolving posed object SDF. An illustration of the entire pipeline is found in \Fig \ref{fig:pipeline}, which we refer the reader back to throughout this section. 

\subsection{Task definition}
\label{ssec:task_def}

\ours incrementally builds an \textit{object model}, simultaneously optimizing for the object SDF network's weights $\theta$ and its corresponding pose $\mathbf{x}_t$ at the current timestep $t$. For object exploration, we use a proprioception-driven policy $\mathbf{\pi}_t$ that executes the optimal action to achieve stable rotation. The input stream of all sensors $\mathcal{S}$ consists of the following (left column of \Fig \ref{fig:pipeline}):
\begin{itemize}
    \setlength{\itemsep}{-5pt} %
    \item \textbf{RGB-D vision}: image $I_t^c$ and depth $D_t^c$ from calibrated camera $c \in \mathcal{S}$
    \item \textbf{RGB touch}: images $I_t^s$ from four DIGITs~\cite{lambeta2020digit}; $s \in \{d_{\text{index}}, d_{\text{middle}}, d_{\text{ring}}, d_{\text{thumb}} \} \in \mathcal{S}$
    \item \textbf{Proprioception}: joint-angles $\mathbf{q}_t$ from robot encoders.
\end{itemize}

\subsection{Robot hardware and simulation}
\label{ssec:setup}

The Allegro hand~\cite{allegro2023} is retrofit with four DIGIT vision-based tactile sensors~\cite{lambeta2020digit}, at each of the distal ends. The DIGIT produces a $240\!\times\!320$ RGB image of the physical interaction at $30\,\text{Hz}$. The Allegro publishes 16D joint-angles so as to situate the tactile sensors with respect to the base frame. The hand is rigidly mounted on a Franka Panda arm, with an Intel D435 RGB-D camera placed at approximately $35\,\text{cm}$ from it. The camera extrinsics are computed with respect to the base frame of the Allegro through ArUco~\cite{garrido2014automatic} hand-eye calibration. For our vision pseudo-ground-truth we use three such cameras in the workspace (\Fig \ref{fig:robot_cell}), jointly calibrated via Kalibr~\cite{furgale2013unified}, to achieve $\approx1\,\text{px}$ reprojection error. Our simulator replicates the real-world setup: a combination of the IsaacGym physics simulator~\cite{makoviychuk2021isaac} with the TACTO touch renderer~\cite{wang2022tacto}. In this case, we can record and store the true ground-truth object pose directly from IsaacGym.

\subsection{\dataset: a visuo-tactile perception dataset}
\label{ssec:feelsight}
\begin{figure}[t]
    \thisfloatpagestyle{empty}
    \centering
    \includegraphics[width=\columnwidth,keepaspectratio]{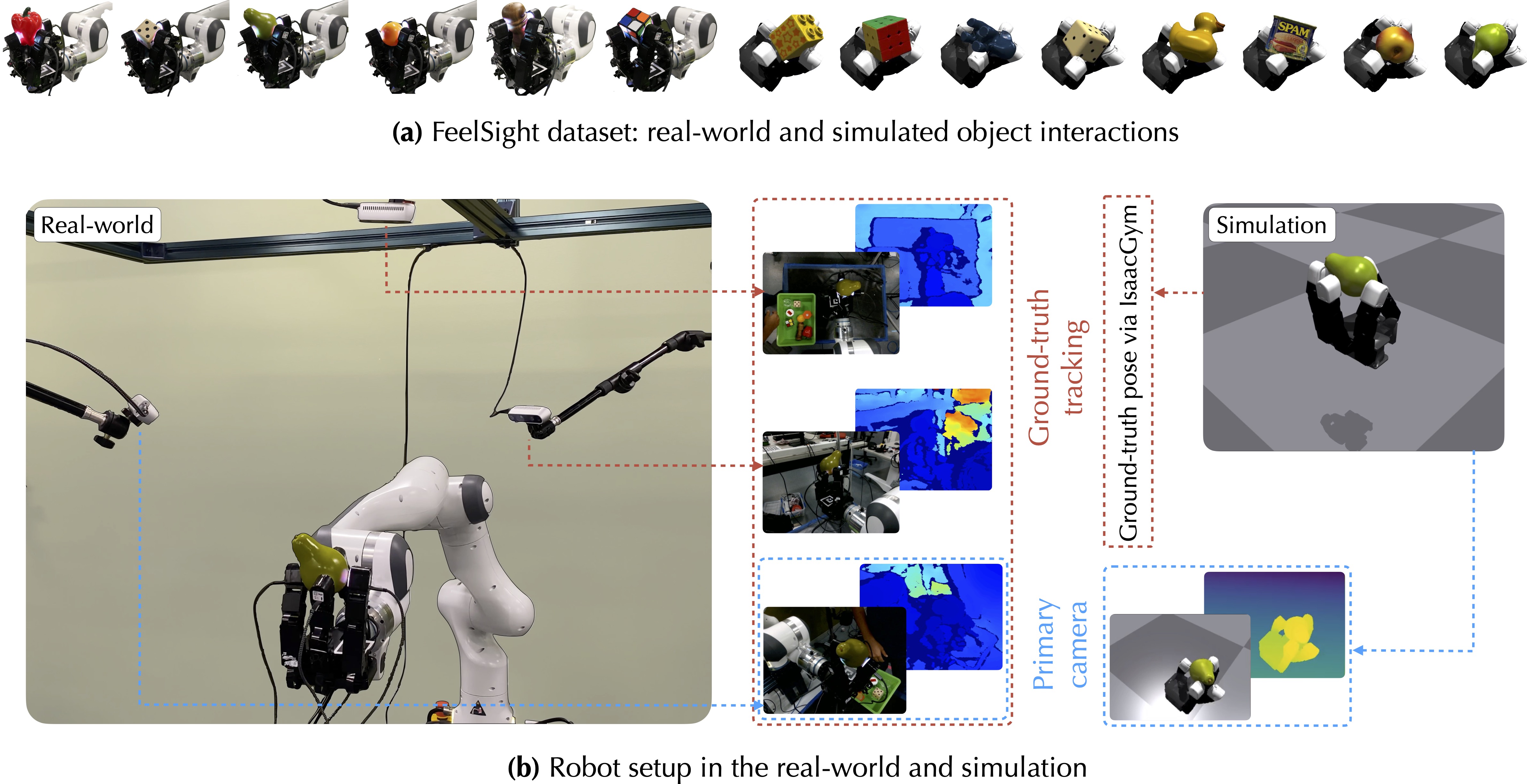}
    \caption{\textbf{Robot setup in the real-world and simulation.} \textbf{(a)} We capture diverse visuo-tactile interactions across different object categories in the real-world and physics simulation. \textbf{(b)} The robot cell is made up of three realsense RGB-D cameras, an Allegro robot hand mounted on a Franka Panda, and four DIGIT tactile sensors. All real-world results use the primary camera and DIGIT sensing, while the additional cameras are fused for our ground-truth pose tracking. In simulation, we use an identical primary camera in IsaacGym with touch simulated in TACTO. The simulator provides ground-truth object pose, so multi-camera tracking is not necessary.}
    \label{fig:robot_cell}
\end{figure}
Visuo-tactile perception lacks a standardized benchmark or dataset that has driven progress in adjacent fields like visual tracking~\cite{hodan2018bop}, SLAM~\cite{geiger2013vision}, and reinforcement learning~\cite{james2020rlbench}. Towards this, we introduce our \dataset dataset for visuo-tactile manipulation. We use the in-hand rotation policy (\Sec \ref{ssec:hora}) to collect vision, touch, and proprioception for $30$ seconds per trial. 

When we encounter a novel object, we tend to twirl it in our hand to get a better look from different views, and regrasp it from different angles. The equivalent for a multi-fingered hand, in-hand rotation, is an ideal choice for the interactive perception problem. We adopt the method of Qi~\etal~\cite{qi2023hand} where they train a proprioception-based policy in simulation, and directly transfer it to the real-world. Recent work has further shown in-hand object rotation using touch and proprioceptive history~\cite{qi2023general, yin2023rotating}, however our simpler abstraction proves sufficient for this task.  In our experiments, the rotation policy $\mathbf{\pi}_t$ sends commands to the robot hand at $20\,\text{Hz}$ via the ROS Allegro controller. This achieves stable rotation of novel objects and interesting visuo-tactile stimuli; for further details refer to \Sec \ref{ssec:hora}. 

The dataset has $5$ in-hand rotation trials each of $6$ objects in the real-world and $8$ objects in simulation; a total $35$ minutes of interaction. As explained in \Fig \ref{fig:robot_cell}, we record a pseudo-ground-truth in the real-world, and exact ground-truth poses in simulation. We ensure diversity in the class of objects: they vary in geometry and size from $6$-$18\,\text{cm}$ in diagonal length. Ground-truth meshes of each object are obtained with the Revopoint 3D scanner~\cite{revopint2023}, which uses dual-camera infrared for $\approx 0.05\,\text{mm}$ scan accuracy.  Additionally, the the simulated experiments have ground-truth meshes from the YCB~\cite{calli2017yale} and ContactDB~\cite{brahmbhatt2019contactgrasp} datasets.

\subsection{Method overview and key insights}
\label{ssec:overview}

\textbf{Object model (\Sec \ref{ssec:object_model}):} We represent the object SDF as a neural network with weights $\theta$, whose output is transformed by the current object pose $\mathbf{x_t}$. This continuous function $F_{\mathbf{x}_t}^\theta(\mathbf{p}) : \mathbb{R}^3 \rightarrow \mathbb{R}$ maps a 3D coordinate $\mathbf{p}$ to a scalar signed-distance from the object's closest surface. Online updates are decomposed into alternating steps between refining the weights of the neural SDF $\theta$, and optimizing the object pose $\mathbf{x_t}$. Our bespoke object model is a representation of both the pose and object geometry over time. 
\ifarxiv 
\vspace{0.5em}
\fi 
\\\textbf{Frontend (\Sec \ref{sssec:segmented_vision}, \ref{sssec:dpt_tactile}):} Given the RGB-D, RGB, and proprioception inputs, our frontend returns segmented depth measurements compatible with our backend optimizer. These modules are pre-trained with a large corpus of data.
\ifarxiv 
\vspace{0.5em}
\fi 
\\\textbf{Shape optimizer (\Sec \ref{ssec:shape_optimizer}):} Takes in frontend output and optimizes for $\theta$ at fixed object pose $\mathbf{\bar{x}_t}$ via gradient descent~\cite{muller2022instant}. Each shape iteration results in improved object SDF $F_{\mathbf{\bar{x}}_t}^\theta$.
\ifarxiv 
\vspace{0.5em}
\fi 
\\\textbf{Pose optimizer (\Sec \ref{ssec:pose_optimizer}):} Builds and optimizes an object pose-graph~\cite{pineda2022theseus} for $\mathbf{x}_t$ given fixed network weights $\bar{\theta}$. Every pose iteration spatially aligns the evolving object SDF with the current set of frontend output.

\subsubsection{Insight 1: \ours is a posed neural field}  
The object model $F_{\mathbf{x_t}}^\theta$ is estimated by a chicken-and-egg optimization of both the instant-NGP weights $\theta$, and the object pose $\mathbf{x}_t$. Prior work has estimated the pose of a sensor in fixed neural field, either by freezing the network weights~\cite{yen2021inerf, lin2021barf}, or joint-optimization~\cite{sucar2021imap, zhu2022nice, rosinol2022nerf}. In our case, robot kinematics gives us the pose of the touch sensors, and extrinsics give us the pose of the camera. So, we instead \textit{flip} this paradigm to estimate the pose of the neural field with respect to known-pose sensors. 

\subsubsection{Insight 2: Touch is vision, albeit local} 
We extend neural fields to directly incorporate touch just as it would vision. Our key insight is that vision-based touch can be approximated as a perspective camera model in tactile simulators like TACTO~\cite{wang2022tacto}. There are, however, differences that must be accounted for in image formation \textbf{(i)} vision-based tactile sensor impose their own color and illumination to the scene, which makes it hard to get reliable visual cues, \textbf{(ii)} a tactile image stream has considerably smaller metric field-of-view and depth-range is usually in centimeters rather than meters, \textbf{(iii)} tactile images have depth discontinuities along \textit{all} non-contact regions, while natural images only encounter them along occlusion boundaries. Our method adapts each of these by \textbf{(i)} consistently using depth rather than color for optimization, \textbf{(ii)} sampling at different scales (centimeter v.s. meter) based on sensing source, \textbf{(iii)} sampling only surface points for touch, but both free-space and surface points for vision. More details are described in \Sec \ref{ssec:shape_optimizer}. After accounting for these differences, we can sample touch consistent with vision, giving us a rich perspective of the object. 

\subsection{Object model}
\label{ssec:object_model}

Our object model is depicted in the right column of \Fig \ref{fig:pipeline}. In general, a neural SDF~\cite{ortiz2022isdf, azinovic2022neural, muller2022instant} represents 3D surfaces as the zero level-set of a learnable function $F(\mathbf{p}) : \mathbb{R}^3 \rightarrow \mathbb{R}$. The scalar field's sign indicates if any query point $\mathbf{p}$ in the volume is inside (negative), outside (positive) or on ($\approx$ 0) the reconstructed surface. $\mathbf{p}$ is first \textit{positionally-encoded}~\cite{tancik2020fourier} into a higher-dimensional space, an important routine that helps networks better approximate high-frequency surfaces.  This is followed by a multi-layer perceptron (MLP) that fits the encoding to a scalar field. Typically, this network is optimized with depth samples from a camera of known intrinsics, and annotated poses from structure-from-motion~\cite{schonberger2016structure}. 

A neural SDF is more compact than the more popular neural radiance fields~\cite{mildenhall2021nerf}, as they do not model color and appearance properties of the scene.  This is sufficient for manipulation, as we care more about estimating geometry than generating novel-views. Recently, instant-NGP~\cite{muller2022instant} demonstrated a learnable multiresolution hash table as a positional encoding that greatly accelerates SDF optimization with small MLP backbones. This has been successfully leveraged for real-time SLAM in indoor scene~\cite{rosinol2022nerf}.

In our work, $F_{\mathbf{x}_t}^\theta$ represents the neural SDF of the object at a given pose $\mathbf{x}_t$. While $\mathbf{x}_t$ is initialized to be between the robot fingers, $\theta$ is randomly initialized. Both shape and pose are estimated via alternating optimization, which emulating the paradigm of tracking and mapping that has found great success in robot vision~\cite{cadena2016past}. The model is fully-differentiable, can be queried arbitrarily in 3D space, and easily extensible to color, latent physics, and other properties.

\subsection{Frontend}
\label{ssec:frontend}
The frontend processes are shown in the center column of \Fig \ref{fig:pipeline}. Its function is to robustly extract depth measurements from raw vision and touch sensing. Depth is available as-is in an RGB-D camera, but the challenge is to robustly segment out object depth pixels in heavily-occluded interactions. Towards this, we introduce a kinematics-aware segmentation strategy using powerful vision foundation models~\cite{kirillov2023segment} (\Sec \ref{sssec:segmented_vision}). Estimating depth from vision-based touch is an open research problem~\cite{bauza2019tactile, wang2021gelsight, sodhi2021patchgraph, ambrus2021monocular, suresh2022shapemap} where millimeter precision and generalization across sensors is important. Towards this, we present a transformer architecture that accurately predicts DIGIT contact patches from inputs images (\Sec \ref{sssec:dpt_tactile}). Unlike our backend that is optimized online, the frontend networks are pre-trained from a large corpus of data. The output of our frontend is a segmented depth image $\hat{D}_t^s$ for each sensor $s \in \mathcal{S}$.
\begin{figure}[ht!]
    \thisfloatpagestyle{empty}
    \centering
    \includegraphics[width=\columnwidth,keepaspectratio]{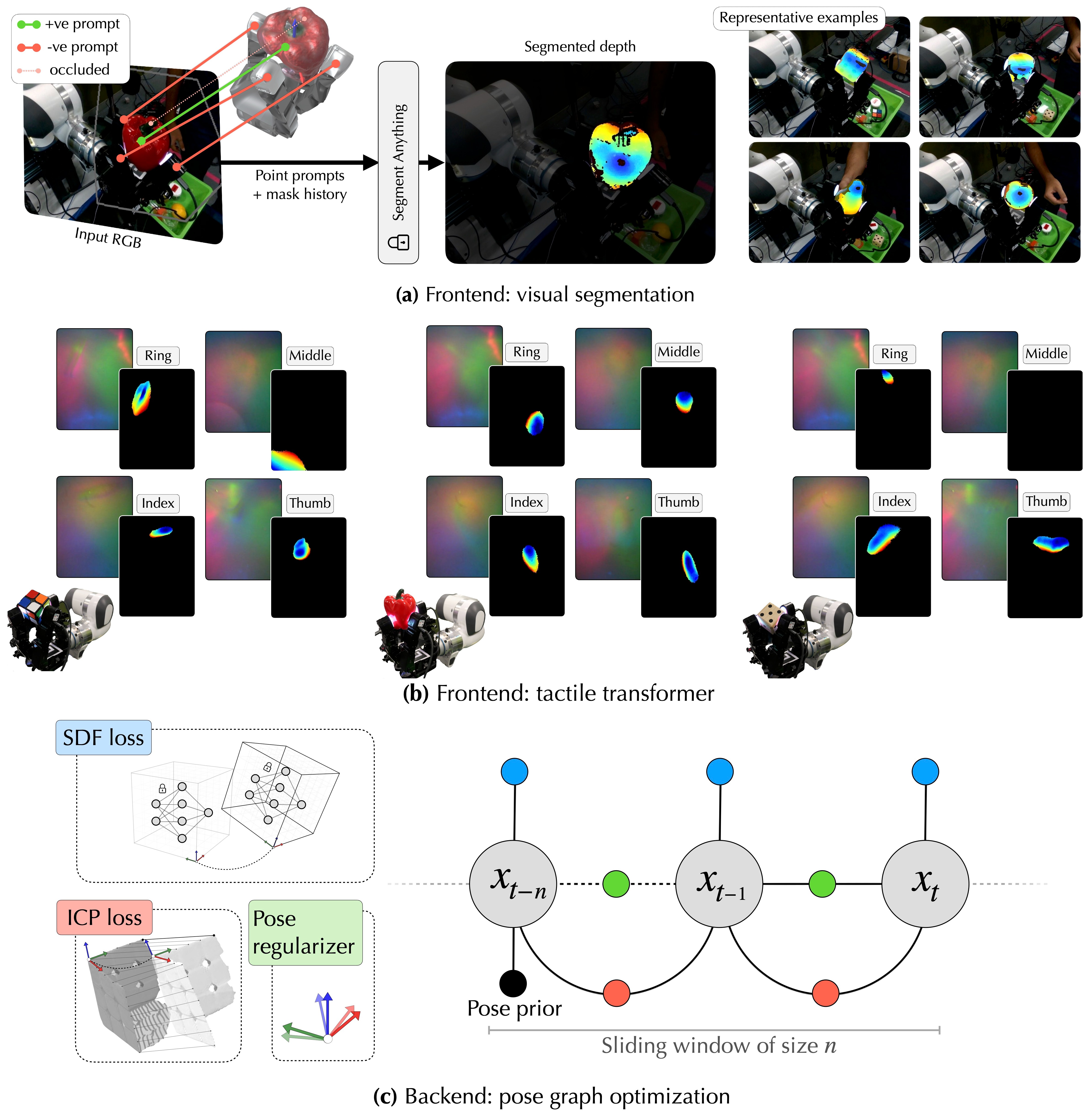}
    \caption{\textbf{Frontend and backend description.} \textbf{(a)} Segment-Anything~\cite{kirillov2023segment} combined with embodied prompts, gives us robust object segmentation. Through reasoning about finger occlusion and object pose with respect to the fingers, we can accurately prompt the segmentation network for robust output masks. \textbf{(b)} Representative examples of the \simtoreal performance of the tactile transformer. Each RGB image is fed through the network to output a predicted depth, along with a contact mask. \textbf{(c)} Our sliding window nonlinear least squares optimizer estimates the object pose $x_t$ from the outputs of the frontend. Each object pose $x_t$ is constrained by the SDF loss, frame-to-frame ICP, and pose regularization to ensure tracking remains stable.}
    \label{fig:frontend_and_backend}
\end{figure}
\subsubsection{Segmented visual depth}
\label{sssec:segmented_vision}
During in-hand manipulation, finger-object occlusion is inevitable and the foreground-background is ambiguous. Robust segmentation of the image stream $I_t^c$ via prompts has successfully been demonstrated by image foundation models, like the Segment Anything Model (SAM)~
\cite{kirillov2023segment}. Trained with a vision transformer (ViT) in the data-rich natural image domain, SAM generalizes to novel scenes for state-of-the-art, zero-shot instance segmentation. 
 
Even with SAM, in-hand object segmentation requires appropriate prompts to guide the pre-trained model. With an embodied agent, we can take advantage of robot kinematics to achieve this. Given our camera $c$ with known projection operation  $\Pi^c$, we can obtain any 3D point $\mathbf{p}$ as a pixel $(u, v) = \Pi^c\left(\mathbf{p}\right)$ on the image $I_t^c$. Our insight is to use the 3D center of grasp and robot kinematics as prompts for SAM (refer to \Sec \ref{ssec:additional}). This makes the reasonable assumption that the object exists between the robot's fingers, which is almost always the case. In \Fig~\ref{fig:frontend_and_backend} (a) we visualize the segmentation on real-world images, alongside the SAM prompts. In our experiments we use the ViT-L model with $308\,\!\text{M}$ parameters. While this achieves a speed of around $4$Hz, in practice, we can use efficient segmentation models~\cite{mobile_sam} for speeds up to $40$Hz. 

\subsubsection{Tactile transformer}
\label{sssec:dpt_tactile}
In contrast, vision-based touch images are out-of-distribution from images SAM is typically trained on, and does not directly provide depth either. The embedded camera perceives an illuminated gelpad, and contact depth is either obtained via photometric stereo~\cite{yuan2017gelsight}, or supervised learning~\cite{bauza2019tactile, wang2021gelsight, sodhi2021patchgraph, ambrus2021monocular, suresh2022shapemap}. Existing touch-to-depth relies on convolution, however recent work has shown the benefit of a ViT for dense depth prediction~\cite{ranftl2021vision} in natural images. We train a \textit{tactile transformer} for predicting contact depth from vision-based touch to generalize across \textit{multiple} real-world DIGIT sensors. 

The architecture is trained entirely in tactile simulation, using weights initialized from a pre-trained image-to-depth model~\cite{ranftl2021vision}. The tactile transformer represents the inverse sensor model $\mathbf{\Omega}: I_t^s \mapsto \hat{D}_t^s$ where $s \in \{d_{\text{index}}, d_{\text{middle}}, d_{\text{ring}}, d_{\text{thumb}} \} \in \mathcal{S}$. This architecture is based on the dense vision transformer~\cite{ranftl2021vision} and is lightweight (21.7M parameters) compared to its fully-convolution counterparts~\cite{suresh2022midastouch}. 

Similar to prior work~\cite{suresh2022shapemap, suresh2022midastouch}, we generate a large corpus of tactile images and paired ground-truth depthmaps in the optical touch simulator TACTO~\cite{wang2022tacto}. We collect 10K random tactile interactions each on the surface of 40 unique YCB objects~\cite{calli2017yale}. For \simtoreal transfer we augment the data with randomization in sensor LED lighting, indentation depth, and pixel noise. In TACTO, image realism is achieved by compositing with \textit{template} non-contact images from real-world DIGITs. For details on the training and data, refer to \Sec \ref{ssec:dpt_tactile_training}. 

These augmentations enable generalized performance across our multi-finger platform, where each sensor has differing image characteristics. Our tactile transformer is supervised on mean-square depth reconstruction loss against the ground-truth depthmaps from simulation. Based on the predicted depthmaps, the output is thresholded to mask out non-contact regions. The tactile transformer demonstrates an average prediction error of $0.042\,\text{mm}$ on simulated test set. \Fig \ref{fig:frontend_and_backend} (b) shows \simtoreal performance of the tactile transformer on real-world interactions.

\subsection{Backend: shape and pose optimizer}
\label{ssec:backend}
The backend (right column of \Fig \ref{fig:pipeline}) is responsible for taking in depth and sensor poses from the frontend to build our \textit{object model} online. This alternates between shape (\Sec \ref{ssec:shape_optimizer}) and pose optimization (\Sec \ref{ssec:pose_optimizer}) steps using samples from the visuo-tactile depth stream. Similar to other neural SLAM methods~\cite{ortiz2022isdf}, the modules maintain a bank of \textit{keyframes} over time to generate these samples. Additional implementation details for the backend are found in \Sec \ref{ssec:additional}. 
\subsubsection{Shape optimizer}
\label{ssec:shape_optimizer}

For online estimation it is intractable to optimize $F_{\mathbf{\bar{x}}_t}^\theta$ using \textit{all} input frames as in neural radiance fields~\cite{mildenhall2021nerf}. We opt for an online learning approach~\cite{sucar2021imap, ortiz2022isdf}, which builds a subset of \textit{keyframes} $\mathcal{K}$ on-the-fly to optimize over. The backend must both \textbf{(i)} accept new keyframes based on a criteria, and \textbf{(ii)} replay old keyframes in the optimization to prevent \textit{catastrophic forgetting}~\cite{sucar2021imap}. Each iteration of the shape optimizer replays a batch $k_t \in \mathcal{K}$ of size $10$ per sensor to optimize our network. This includes the latest two frames, and a weighted random sampling of past keyframes based on average rendering loss. 

The initial visuo-tactile frame is automatically added as a keyframe $\mathcal{K}_0 = \{\hat{D}_0^s \ | \ s \in \mathcal{S}\}$, and every subsequent keyframe $\mathcal{K}_t$ is accepted using an information gain metric~\cite{sucar2021imap}. For this, the average rendering loss is computed from the frozen network $F_{\mathbf{\bar{x}}_t}^\theta$ using the given keyframe pose and compared against a threshold $d_{\text{thresh}} = 0.01\,\text{m}$. Finally, if we have not added a keyframe for an interval $t_{\text{max}} = 0.2\,\text{secs}$, we force one to be added. 

\noindent\textbf{{Sampling and SDF loss.}} At each iteration, we sample coordinates in the neural volume from $k_t$ to optimize the neural weights $\theta$. The first step is to sample a batch of pixels $\mathbf{u}_{k_t}$ from $k_t$---a mix of surface and free-space pixels. While surface pixels directly supervise the SDF zero level-set, free-space pixels carve out the neural volume. In our implementation, we sample $50\%$ of camera pixels in free-space, while we only sample surface pixels for touch. Through each pixel $u \in \mathbf{u}_{k_t}$ given their corresponding sensor pose, we project a ray into the neural volume.  Similar to Ortiz~\etal~\cite{ortiz2022isdf}, we sample $P_u$ points per ray, a mix of stratified and surface points.

With these samples, we compute an SDF prediction $\mathbf{\hat{d}}_u$ for each $\hat{D}_t \in k_t$, as the batch distance bound~\cite{ortiz2022isdf}. For each ray, we split the samples into $P_u^{\text{f}}$ and $P_u^{\text{tr}}$ based on $\mathbf{\hat{d}}_u$ lies within the truncation distance $d_{\text{tr}}\!=\!5\,\text{mm}$ from the surface. Our shape loss $\mathcal{L}_{\text{shape}} = \mathcal{L}_{\text{f}} +  w_{\text{tr}}\mathcal{L}_{\text{tr}}$, with $w_{\text{tr}} = 10$, resembles the truncated SDF loss of Azinovi\'{c}~\etal~\cite{azinovic2022neural}: 
\begin{align*} 
 \mathcal{L}_{\text{f}} = \frac{1}{|\mathbf{u}_{k_t}|} \sum\limits_{u \in \mathbf{u}_{k_t}} \frac{1}{|P_u^{\text{f}}|} |F_{\mathbf{\bar{x}}_t}^\theta(P_u^{\text{f}}) - d_{\text{tr}}|
 \qquad \text{and} \qquad
 \mathcal{L}_{\text{tr}} = \frac{1}{|\mathbf{u}_{k_t}|} \sum\limits_{u \in \mathbf{u}_{k_t}} \frac{1}{|P_u^{\text{tr}}|} |F_{\mathbf{\bar{x}}_t}^\theta(P_u^{\text{tr}}) - \mathbf{\hat{d}}_u|
\end{align*} 

\subsubsection{Pose optimizer}
\label{ssec:pose_optimizer}

Before each shape iteration, we use a pose graph~\cite{dellaert2017factor} to refine the object pose $\mathbf{x}_t$ with respect to the frozen neural field $F_{\mathbf{x}_t}^{\bar{\theta}}$. We achieve this by \textit{inverting} the problem to instead optimize for the 6-DoF poses in a sliding window of size $n$. At timestep $t$, if we have accumulated $N$ keyframes, this is represents poses
$\mathcal{X}_t\!=\!{\left(\mathbf{x}_i\right)}_{N\!-\!n \le i \le N}$
and measurements 
$\mathcal{M}_t\!=\!{\left(\hat{D}_i^s \ | \ s\!\in\!\mathcal{S}\right)}_{N\!-\!n \le i \le N}$.
Similar to pose updates in visual SLAM~\cite{yen2021inerf, sucar2021imap, zhu2022nice}, the network weights $\bar{\theta}$ are frozen and we estimate the $\textit{SE}(3)$ poses $\mathcal{X}_t$ instead. 

We formulate the problem as a nonlinear least squares optimization with custom measurement factors in  Theseus~\cite{pineda2022theseus}. While prior work uses gradient descent~\cite{yen2021inerf}, we instead use a second-order Levenberg–Marquardt (LM) solver, which provides faster convergence~\cite{dellaert2017factor}. The pose graph, illustrated in \Fig \ref{fig:frontend_and_backend} (c), solves for:
\begin{align*} 
\hat{\mathcal{X}_t} = \underset{\mathcal{X}_t}{\text{argmin}} \ \mathcal{L_{\text{pose}}}(\mathcal{X}_t  \ | \ \mathcal{M}_t, \bar{\theta}) 
\qquad \text{where} \qquad
\mathcal{L_{\text{pose}}} = w_{\text{sdf}}\mathcal{L}_{\text{sdf}} + w_{\text{reg}}\mathcal{L}_{\text{reg}} + w_{\text{icp}}\mathcal{L}_{\text{icp}}
\end{align*}
\begin{itemize}
    \setlength{\itemsep}{-5pt} %
    \item \textbf{SDF loss $\mathcal{L}_{\text{sdf}}$.} We use the shape loss $\mathcal{L}_{\text{shape}}$, modified such that we sample only about surface points of each ray. This works well for both visual and tactile sensing as we have higher confidence in SDFs about the surface of the object than in free-space. For each depth measurement in $\mathcal{M}_t$, we sample surface points over $M$ rays, and average the SDF loss along each ray. This results in an $M\!\times\!n$ SDF loss, which we use to update the se(3) lie algebra of $\mathcal{X}_t$. We implement a custom Jacobian for this cost function, which is up to 4$\times$ more efficient than PyTorch automatic differentiation. 
    \item \textbf{Pose regularizer $\mathcal{L}_{\text{reg}}$.} We apply a weak regularizer between consecutive keyframe poses in $\mathcal{X}_t$ to ensure the relative pose updates stay well-behaved. This is important for robustness to noisy frontend depth and incorrect segmentations. 
    \item \textbf{ICP loss $\mathcal{L}_{\text{icp}}$.} We further apply iterative closest point (ICP) between the current visuo-tactile pointcloud $\Pi^{-1}(\mathcal{M}_t)$ and previous pointcloud $\Pi^{-1}(\mathcal{M}_{t-1})$. This gives us frame-to-frame constraints in addition to the frame-to-model $\mathcal{L}_{\text{sdf}}$.
\end{itemize}

%% file: sections/acknowledgements.tex
\newpage
\section*{Acknowledgments}\label{sec:acknowledgments}

The authors thank Dhruv Batra, Theophile Gervet, Akshara Rai for feedback on the writing, and Wei Dong, Tess Hellebrekers, Carolina Higuera, Patrick Lancaster, Franziska Meier, Alberto Rodriguez, Akash Sharma, Jessica Yin for helpful discussions on the research. 

\vspace{0.5em}
\noindent \textbf{Author contributions:} \input{sections/contributions}

\vspace{0.5em}
\noindent \textbf{Funding:} Sudharshan Suresh and Haozhi Qi acknowledge funding from Meta, and their work was partially conducted while at FAIR, Meta. Sudharshan Suresh was further partially supported by NSF grant IIS-2008279 while at CMU. Roberto Calandra acknowledge support from the German Research Foundation (DFG, Deutsche Forschungsgemeinschaft) as part of Germany’s Excellence Strategy – EXC 2050/1 – Project ID 390696704 – Cluster of Excellence “Centre for Tactile Internet with Human-in-the-Loop” (CeTI) of Technische Universität Dresden, and from Bundesministerium für Bildung und Forschung (BMBF) and German Academic Exchange Service (DAAD) in project 57616814 (School of Embedded and Composite AI, \href{https://secai.org/}{SECAI}).

\ifarxiv

\else
    \vspace{0.5em}
    \noindent \textbf{Competing interests:} The authors declare that they have no competing interests.
    
    \vspace{0.5em}
    \noindent \textbf{Data and materials availability:} The dataset and code will be made publicly available prior to acceptance, we provide reviewers a pre-release here: \url{\prerelease}. For multimedia, we refer the readers to our project webpage: \url{\webpage}. All the data to validate the paper is available in the main body and supplementary sections. 
\fi 

%% file: sections/contributions.tex
\noindent \textit{Sudharshan Suresh} developed and implemented the core approach including tactile transformer, visual depth segmentation, neural SDF reconstruction, pose-graph optimization, performed full-stack tuning, worked on Allegro and DIGIT integration, TACTO and IsaacGym integration, camera and robot calibration, data collection, ground truth object scans, and live visualizations, conducted evaluations, made visuals, and wrote the paper.

\noindent \textit{Haozhi Qi} designed and implemented in-hand object rotation policies and sim-to-real policy transfer, helped with Allegro and DIGIT integration, TACTO and IsaacGym integration, data collection, did code reviews and bug fixes, and helped edit the paper.

\noindent \textit{Tingfan Wu} coordinated hardware and software systems integration, performed profiling of software stack, helped with Allegro and DIGIT integration, camera and robot calibration, ground truth object scans, and advised on evaluations.

\noindent \textit{Taosha Fan} designed and implemented forward kinematics, helped implement visual depth segmentation, pose-graph cost functions and optimization, software systems integration, and advised on evaluations.

\noindent \textit{Luis Pineda} implemented workflow for cluster deployment, streamlined development workflow, helped with modules that use Theseus, did code reviews and bug fixes, and advised on evaluations.

\noindent \textit{Mike Lambeta} helped with Allegro and DIGIT integration, TACTO and IsaacGym integration, and hardware systems integrations.

\noindent \textit{Jitendra Malik} advised on the project, gave feedback on approach, evaluations, and the paper.

\noindent \textit{Mrinal Kalakrishnan} advised on the project, managed and supported researchers, gave feedback on approach, evaluations, and the paper.

\noindent \textit{Roberto Calandra} advised on the project, helped with Allegro and DIGIT integration, TACTO and IsaacGym integration, gave feedback on approach, evaluations, and the paper.

\noindent \textit{Michael Kaess} advised on the project, helped design pose-graph optimization, gave feedback on approach, evaluations, and the paper.

\noindent \textit{Joseph Ortiz} advised on the project, co-developed the core approach, implemented volumetric ray sampling, SDF cost function, and 2D live visualizations, helped implement workflow for cluster deployment, streamlined development workflow, did code reviews and bug fixes, gave feedback on evaluations, designed visuals, and edited the paper.

\noindent \textit{Mustafa Mukadam} set the vision and research direction, steered and aligned the team, provided guidance on all aspects of the project including core approach, systems, and evaluations, designed visuals, and edited the paper.

%% file: sections/supplementary.tex
\clearpage
\setcounter{section}{0}
\setcounter{subsection}{0}
\setcounter{figure}{0}
\setcounter{table}{0}
\setcounter{footnote}{0}
\renewcommand{\thesection}{S\arabic{section}}
\renewcommand{\thefigure}{S\arabic{figure}}
\renewcommand{\thetable}{S\arabic{table}}

\makeatletter
\renewcommand\thesubsection{S\@arabic\c@subsection}
\makeatother

\section*{Supplementary materials}
\label{sec:supplementary}

\begin{itemize}
    \setlength\itemsep{-0.3em}
    \item \Sec \ref{ssec:gt_shape_pose} to \ref{ssec:role_touch}
    \item \Fig \ref{fig:ground_truth} to \ref{fig:tactile_features}
    \item Multimedia on our webpage at \url{\webpage}
    \ifarxiv 
    \else
        \item Code and dataset pre-release at \url{\prerelease}
    \fi 
\end{itemize}

\subsection{Ground-truth shape and pose}
\label{ssec:gt_shape_pose}

\begin{figure}[b!]
    \thisfloatpagestyle{empty}
    \centering
    \includegraphics[width=\columnwidth,keepaspectratio]{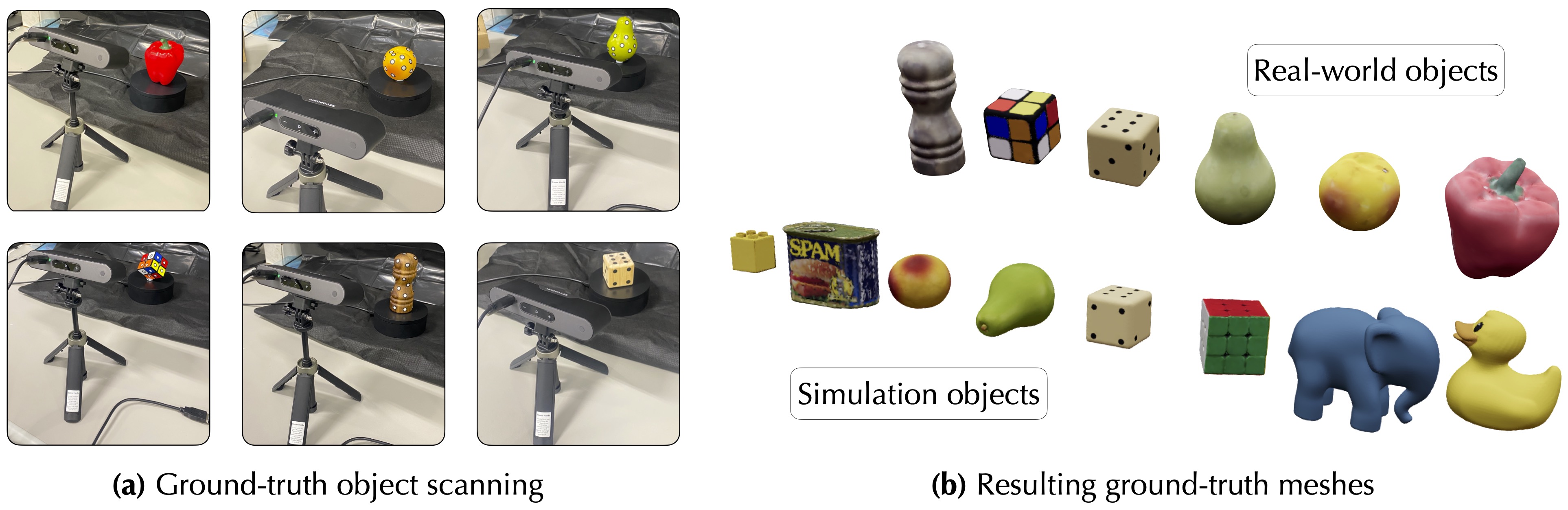}
    \caption{\textbf{Object ground-truth with dual-camera infrared scanner.} \textbf{(a)} Objects are placed on a turntable and scanned, followed by post-processing to ensure complete, accurate meshes. \textbf{(b)} Meshes visualized for the real and simulated \dataset objects.}
    \label{fig:ground_truth}
\end{figure}

\noindent\textbf{Ground-truth object scans.} Our results in \Sec \ref{sec:results} require ground-truth object shape to compare against. For this, we use a commercial dual-camera infrared scanner, the Revopoint POP 3~\cite{revopint2023}. The hardware can scan objects from a close range with a minimum precision of $0.05\,\text{mm}$. Each real-world object is placed on a turntable and scanned while rotating about its axis (\Fig \ref{fig:ground_truth} (a)). For object's that lack texture, an artificial dot pattern is tracked by adding stickers. After generating the scans, we perform hole-filling for unseen regions of the object, like the bottom. \Fig \ref{fig:ground_truth} (b) shows all the scanned meshes---a few meshes are directly sourced from the YCB~\cite{calli2017yale} and ContactDB~\cite{brahmbhatt2019contactgrasp} datasets. 

\begin{figure}[t]
    \thisfloatpagestyle{empty}
    \centering
    \includegraphics[width=\columnwidth,keepaspectratio]{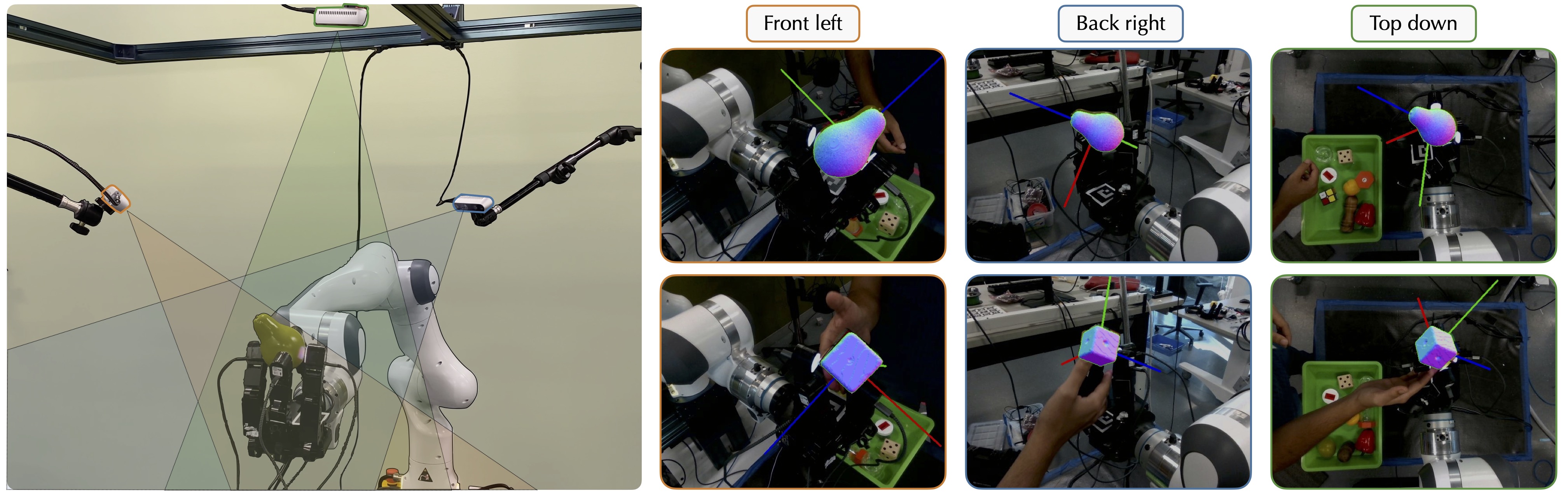}
    \caption{\textbf{Robot cell for pseudo-ground-truth tracking.} Each of the three camera's captures a different field-of-view of the interaction \textbf{(left)}. For a pseudo-ground-truth, we pass the RGB-D stream from all three cameras into our pipeline, with known shape obtained from scanning. The output pose tracking \textbf{(right)} represents the ground-truth we compare to in the real-world results.}
    \label{fig:three_cam}
\end{figure}

\begin{figure}[t!]
    \thisfloatpagestyle{empty}
    \centering
    \includegraphics[width=\columnwidth,keepaspectratio]{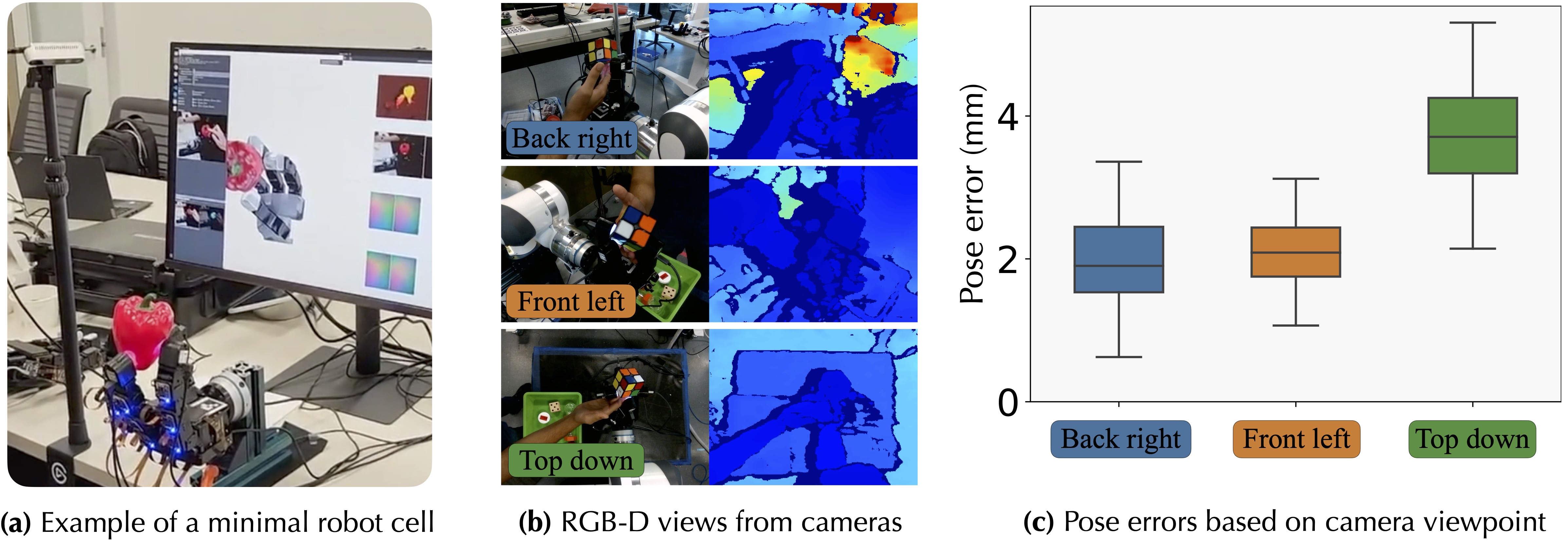}
    \caption{\textbf{(a)} As a proof-of-concept, we assembled a minimal robot cell for live demonstrations of our method with one RGB-D camera and the robot policy deployed at $2$Hz. \textbf{(b)} The three different RGB-D camera viewpoints in our full robot cell used to collect \dataset evaluation dataset. \textbf{(c)} Average pose error for known shape experiments based on camera viewpoint. We see that while the front and back cameras perform comparably, there are larger errors in the top-down camera as it is further away.}
    \label{fig:camera_viewpoint}
\end{figure}

\noindent\textbf{Pseudo ground-truth pose.} In the real-world, we pass three RGB-D cameras as input into our pose tracking pipeline to use as a pseudo ground-truth estimate. This consists of three unique cameras (\textit{front left, back right, top down}) with complementary but overlapping fields-of-view (\Fig \ref{fig:three_cam} and \Fig \ref{fig:camera_viewpoint} (b)). With this broad perspective of the scene, known shape from ground-truth scans, and the tracker running at $0.5$Hz, we can obtain an accurate estimation of object pose at each timestep. 

\subsection{Tactile transformer: data and training}
\label{ssec:dpt_tactile_training}
\begin{figure}[t]
    \thisfloatpagestyle{empty}
    \centering
    \includegraphics[width=0.95\columnwidth,keepaspectratio]{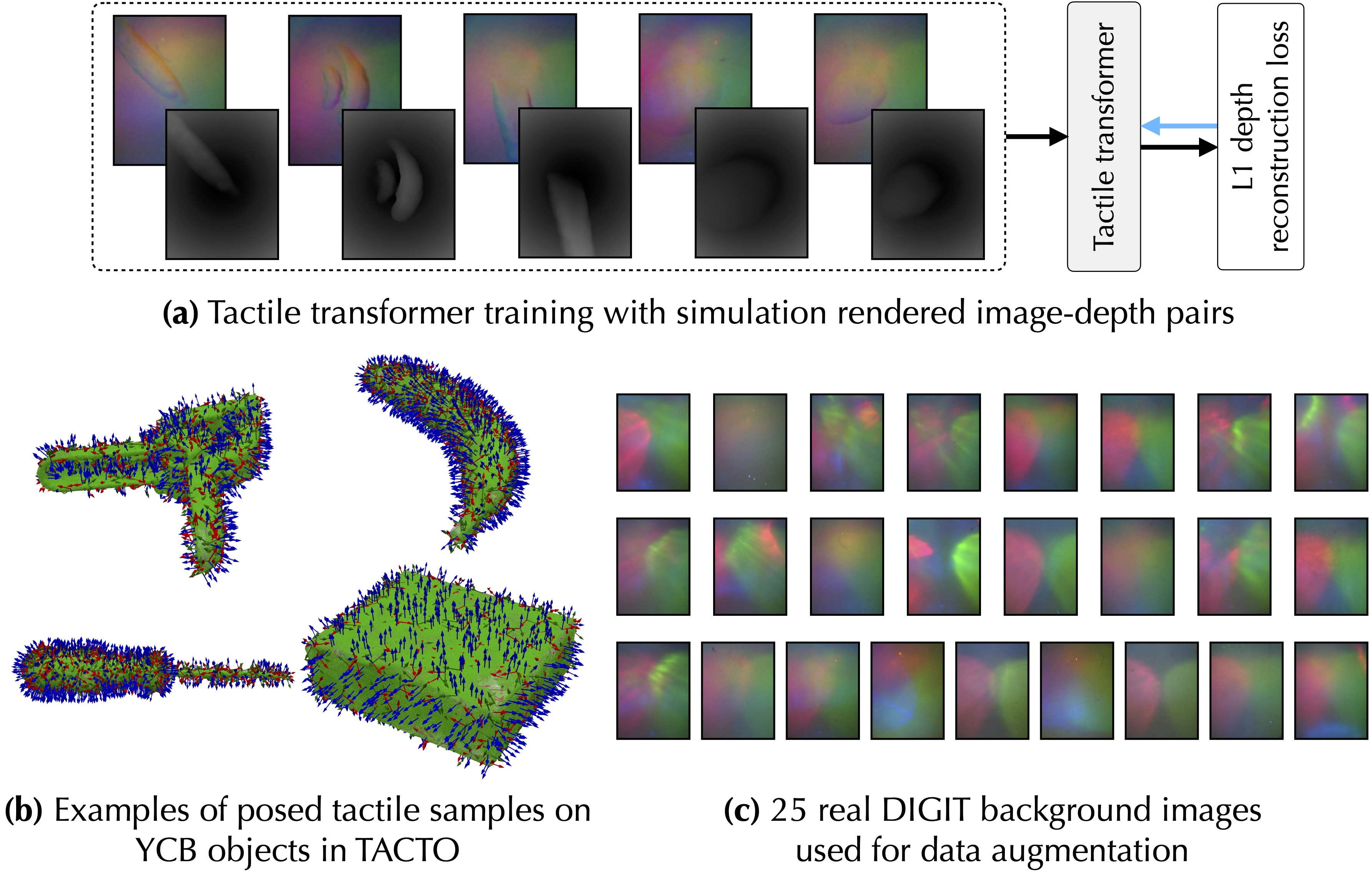}
    \caption{\textbf{Our tactile transformer is trained in simulation with real-world augmentation.} \textbf{(a)} The tactile transformer is supervised from paired RGB-depth images rendered in TACTO~\cite{wang2022tacto}. \textbf{(b)} Each of these samples are generated from dense, random interactions with 40 different YCB objects. \textbf{(c)} In our training, we augment the data with background images collected from $25$ unique DIGIT sensors~\cite{lambeta2020digit}.}
    \label{fig:tactile_transformer_training}
\end{figure}
\noindent\textbf{Model architecture.} Our model architecture is based on a monocular depth network, the dense prediction transformer (DPT)~\cite{ranftl2021vision}. It comprises of a vision transformer (ViT) backbone that outputs bag-of-words features at different resolutions, finally combined into a dense prediction via a convolutional decoder. Compared to fully-convolutional methods, DPT has a global receptive field and the resulting embedding does not explicitly down-sample the image. 

\vspace{1em}

\begin{figure}[t]
    \thisfloatpagestyle{empty}
    \centering
    \includegraphics[width=\columnwidth,keepaspectratio]{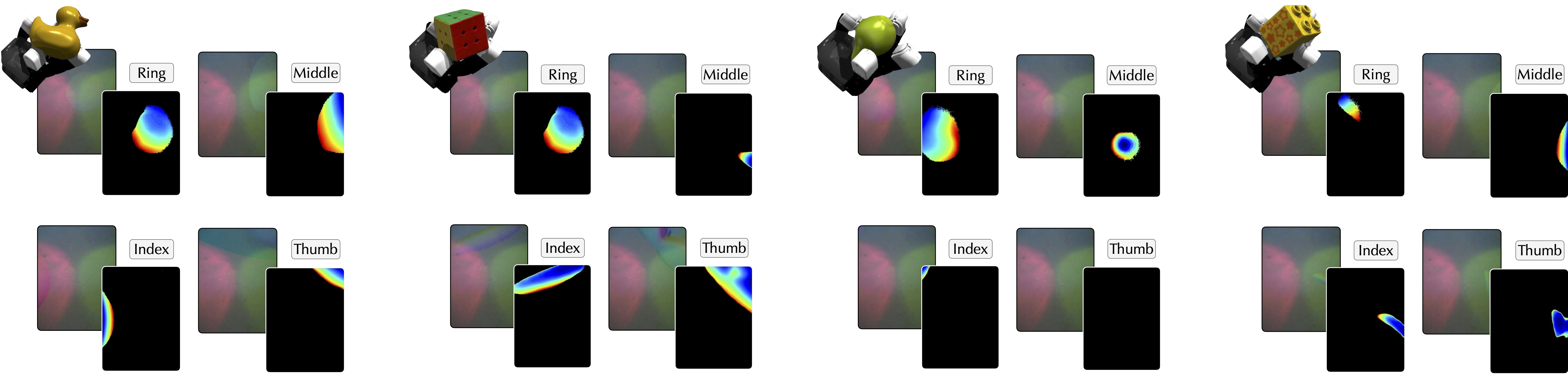}
    \caption{\textbf{Image to depth predictions by the tactile transformer on simulated contacts.} Our tactile transformer shows good performance in simulated interactions, capturing both large contact patches, as well as smaller edge features. These objects are unseen during training---as highlighted in \Sec \ref{sssec:dpt_tactile}, we demonstrate an average prediction error of $0.042\,\text{mm}$ on simulated test images.}
    \label{fig:tactile_transformer_sim}
\end{figure}

\noindent\textbf{Training and loss metric.}
Our image-to-depth training dataset comprises of 10K simulated tactile interactions each on the surface of $40$ YCB objects. We illustrate examples of the interactions in \Fig \ref{fig:tactile_transformer_training} (b). We use the ADAM optimizer with momentum and a batch size of $100$, trained with mean-square depth reconstruction loss (\Fig \ref{fig:tactile_transformer_training} (a)). We start with a pre-trained small ViT~\cite{dosovitskiy2020image}, with an embedding dimension of $384$ patch size of $16$. The dataloader splits the train, test, and validation data into 60\%, 20\%, and 20\% respectively. To supplement our results in \Sec \ref{sssec:dpt_tactile}, we visualize additional simulation results in \Fig \ref{fig:tactile_transformer_sim}. 

\vspace{1em}

\noindent\textbf{Data augmentation.}
An important aspect of generalization and \simtoreal transfer is the augmentation applied during data collection and training. These include: 
\vspace{-0.5em}
\begin{itemize}
    \setlength\itemsep{-0.5em}
    \item \textbf{Real-world backgrounds.} We compose simulated renderings with real-world background images, collected from $25$ different DIGIT sensors. These are shown in \Fig \ref{ssec:dpt_tactile_training} (c). 
    \item \textbf{Pose variations.} Before rendering a sensor pose, we apply noise in rotation/translation and sensing normal direction. Additionally, we randomly vary the distance of penetration into the object surface. 
    \item \textbf{Sensor lighting.} We randomize position, direction and intensity of the three DIGIT LEDs.
    \item \textbf{Sensor pixel noise.} We add Gaussian noise to RGB data, with a standard deviation of $7$px.
    \item \textbf{Standard transforms.} Randomized horizontal flipping, cropping, and rotations of the tactile images.
\end{itemize}

\subsection{Additional implementation details}
\label{ssec:additional}

\textbf{Segmented visual depth.} As discussed in \Sec \ref{sssec:segmented_vision}, we use the 3D center of grasp, by computing the centroid of the end-effectors as a positive point prompt for SAM. However, in practice, this prompt alone doesn't suffice. First, the robot fingers frequently appear in these segmentations, which is misleading to our shape optimizer. This is solved by adding negative point prompts to fingertip pixels that we obtain by projecting the forward kinematics results. We first verify if the fingertips are unoccluded by the object, which we do by comparing against the current rendered object model. Second, SAM tends to over segment objects with distinct parts (\eg different faces of the Rubik's cube). In case of these ambiguities, SAM outputs multiple masks, at different spatial scales. We apply a final pruning step to find the mask prediction closest to the average mask area we typically observe in simulation.

\textbf{{Shape optimizer.}} The neural field is optimized via
Adam~\cite{kingma2014adam} with learning rate of $2\text{e-}4$ and weight decay of $1\text{e-}6$. Instant-NGP uses a hash table of size $2^{19}$ for positional encoding, followed by a 3-layer MLP with $64$ dimensional width. We use a uniform random weights $\theta_{\text{init}}$ and initialize the SDF by running $500$ shape iterations using the first keyframe $\mathcal{K}_0$. 

For evaluating the neural field we freeze the network and query a $200^3$ feature grid. The feature grid's extents are defined as a bounding box of $15\,\text{cm}$ side, centered at the object's initial pose $\mathbf{x}_0$. When training, we apply a series of bounding-box checks post hoc, to eliminate any ray samples $P_u$ found outside this bounding box. Mesh visualizations (\Fig \ref{fig:slam_real}) are periodically generated via marching-cubes on the feature grid. We add color to the mesh by averaging the colored object pointcloud with a Gaussian kernel. 

\textbf{{Pose optimizer.}} We use the vectorized $SE(3)$ pose graph optimizer in Theseus~\cite{pineda2022theseus}, with $20$ LM iterations of step size $1.0$. The keyframe window size $n\!=\!3$ and we run $2$ pose iterations for each shape iteration. The weighting factors for each loss are $w_{\text{sdf}}\!=\!0.01$, $w_{\text{reg}}\!=\!0.01$, and $w_{\text{icp}}\!=\!1.0$.

\textbf{{Compute and timings.}} All results in \Sec \ref{sec:results} are generated from playing-back the trials at a publishing rate of $1\,\text{Hz}$. Experimentally, however, we can run the pose optimizer at $10\,\text{Hz}$ and full backend at 5 Hz. \Fig \ref{fig:camera_viewpoint} (a) has a minimal robot setup of an online SLAM system with rotation policy in-the-loop. Experiments are run on an Nvidia GeForce RTX 4090, while the aggregate results are evaluated on a cluster with Nvidia Tesla V100s. 

\subsection{In-hand exploration policy}
\label{ssec:hora}

We first train a policy in simulation with access to an embedding of physical properties such as object position, size, mass, friction, and center-of-mass (denoted as $\mathbf{z}_t$). From the joint-angles $\mathbf{q}_t$ and this embedding $\mathbf{z}_t$, the policy outputs a PD controller target $\mathbf{a}_t\!\in\!\mathbb{R}^{16}$. The policy is trained in parallel simulated environments~\cite{makoviychuk2021isaac} using proximal policy optimization~\cite{schulman2017proximal}. The reward function is a weighted combination of a rotational reward, joint-angle regularizer, torque penalty, and object velocity penalty. The resulting policy can adaptively rotate objects in-hand according to different physical properties. 

During deployment, however, the policy does not have access to these physical properties. The estimator is instead trained to infer $\mathbf{z}_t$ from a history of proprioceptive states, which is in turn fed into the policy $\mathbf{\pi}_t$. A crucial change compared to Qi~\etal~\cite{qi2023hand} is that we train the policy to rotate objects with DIGIT sensors on the distal ends (\Fig \ref{fig:cover}). This results in different gaits, as it \textbf{(i)} relies on finger-object friction instead of gravity, and \textbf{(ii)} learns to maintain contact with the DIGIT gelpads.

\subsection{Additional results}
\label{ssec:add_results}

\begin{figure}[h]
    \thisfloatpagestyle{empty}
    \centering
    \includegraphics[width=0.95\columnwidth,keepaspectratio]{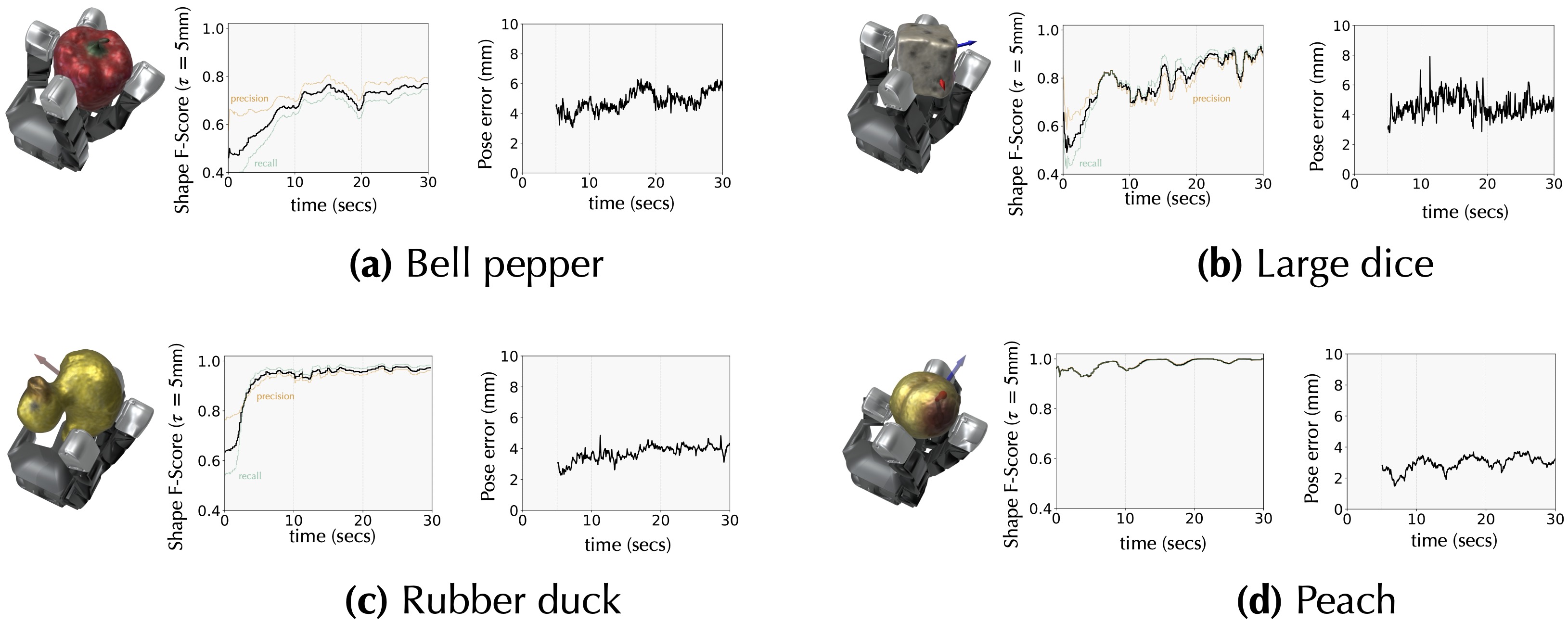}
    \caption{\textbf{Shape and pose metrics over time for in-hand SLAM.} Here, we plot the time-varying metrics for experiments visualized in \Fig \ref{fig:slam_real}. First, we note the gradual increase in F-score over time with further coverage. Additionally, we have bounded pose drift over time---for each experiment we omit the first five seconds as the metric is ill-defined then.}
    \label{fig:metrics_over_time}
\end{figure}

\noindent\textbf{Shape and pose metrics over time.} In \Fig \ref{fig:metrics_over_time}, we plot these metrics for each of the experiments in \Fig \ref{fig:slam_real}, instead against $0\!-\!30\,\text{sec}$ timesteps. For shape, we observe gradual convergence to an asymptote close to $1.0$, indicating evolution of both shape completion and refinement over time. Also visualized here is the precision and recall metrics over time, whose harmonic mean represents the F-score. For pose, we observe stable drift over time, indicating the estimated object pose lies close to the ground-truth estimate. 

\vspace{1em}

\noindent\textbf{Effect of camera viewpoint in the real-world.} In \Sec \ref{ssec:occlusion}, we establish the relationship between occlusion/sensing noise and pose error. Here, we run additional experiments, on a limited set of viewpoints in the real-world. \Fig \ref{fig:camera_viewpoint} (b) shows the RGB-D data from three cameras \textit{front left, back right, top down}, at distances of $27\,\text{cm}$, $28\,\text{cm}$, and $49\,\text{cm}$ respectively from the robot. We run our vision-only pose tracker with known shape using each of three cameras over all $5$ Rubik's cube rotation experiments and plot the average metrics in \Fig \ref{fig:camera_viewpoint} (c). We observe that the \textit{front left} and \textit{back right} viewpoints result in lowest average pose error due to their closer proximity. The \textit{top down} camera gives less reliable depth measurements and segmentation output, leading to almost $2$x greater pose error. 

\vspace{1em}
\begin{figure}[t]
    \thisfloatpagestyle{empty}
    \centering
    \includegraphics[width=\columnwidth,keepaspectratio]{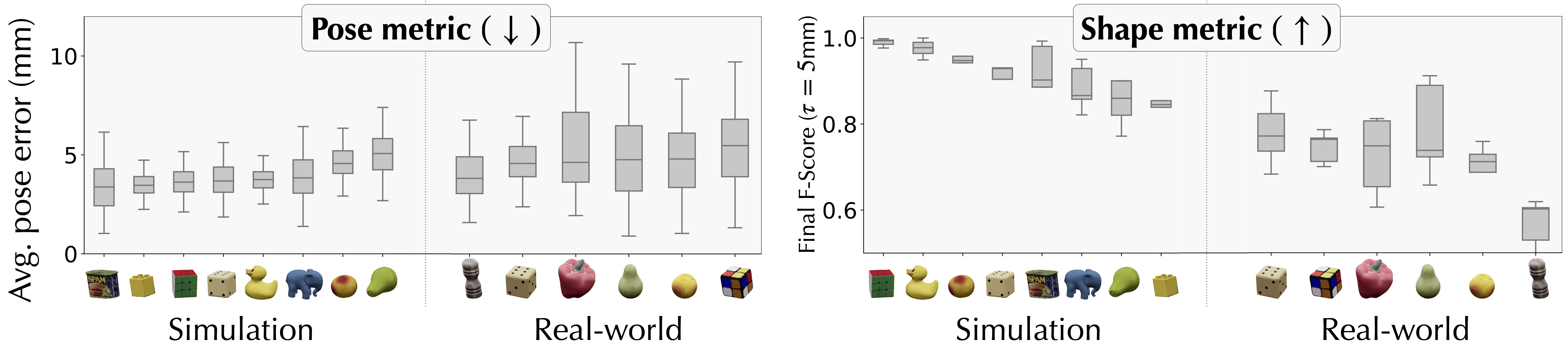}
    \caption{Pose \textbf{(left)} and shape \textbf{(right)} metrics for each object class, sorted in best-to-worst performance.}
    \label{fig:per_object_slam_error}
\end{figure}
\noindent\textbf{Class-specific metrics.} In \Fig \ref{fig:per_object_slam_error}, we present our metrics for the SLAM results in \Sec \ref{ssec:neural_slam}, dividing based on object class. This helps us make some assessments on how object geometry and scale can affect our results. Some observations include: 
\vspace{-0.5em}
\begin{itemize}
    \setlength\itemsep{-0.5em}
    \item \textbf{Object symmetry.} Objects with symmetries about their rotation axis are challenging for our depth-based estimator. This leads to higher pose errors for the \textit{peach} and \textit{pear}, for example. 
    \item \textbf{Object visibility.} Partial visibility of the large objects, such as the \textit{pepper grinder}, affect the completeness of the reconstructions. Touch in this case is not advantageous since the finger gait does not span the length of the object to provide coverage. 
    \item \textbf{Object scale.} Smaller-sized objects, such as the \textit{peach}, may demonstrate better shape metrics as their scale is closer to the F-score threshold of $5\,\text{mm}$.
\end{itemize}

\subsection{Additional visualizations}
\label{ssec:add_viz}

\begin{figure}[t]
    \thisfloatpagestyle{empty}
    \centering
    \href{https://suddhu.github.io/neural-feels/video/dataset_zoom.mp4}{\includegraphics[width=\columnwidth,keepaspectratio]{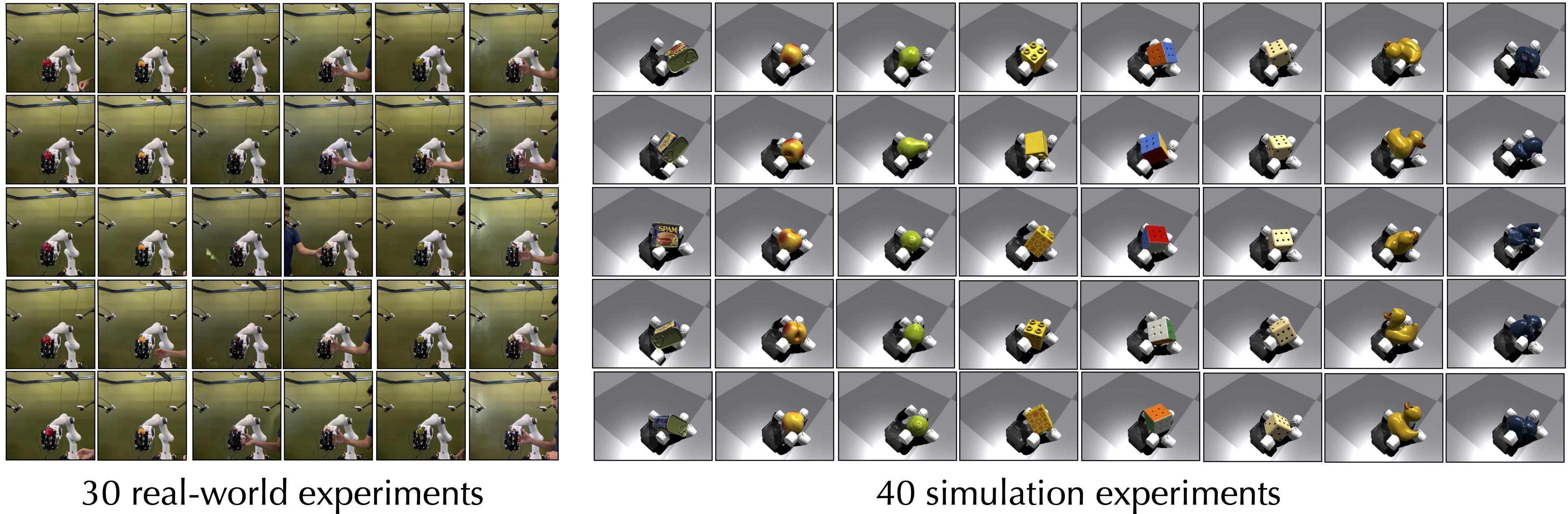}}
    \caption{\textbf{A collage depicting the entirety of the \dataset dataset.} We collect \textbf{(i)} 5 sequences each (row) in the real-world across 6 different objects (column), and \textbf{(ii)} 5 sequences each (row) in simulation across 8 different objects (column).}
    \label{fig:feelsight_full}
\end{figure}

\noindent\textbf{All experiments from the \dataset dataset.} In \Fig \ref{fig:feelsight_full} we illustrate all of the $70$ visuo-tactile experiments that comprise our dataset. While both simulation and real data collection use the proprioception-driven policy~\cite{qi2023hand}, the policy generalizes better in simulation across the class of objects. Some objects in the real-world require a human-in-the-loop to assist with in-hand rotation; \eg supporting cube-shaped objects from the bottom to occasionally prevent falling out of hand.  

\begin{figure}[h]
    \thisfloatpagestyle{empty}
    \centering
    \includegraphics[width=\columnwidth,keepaspectratio]{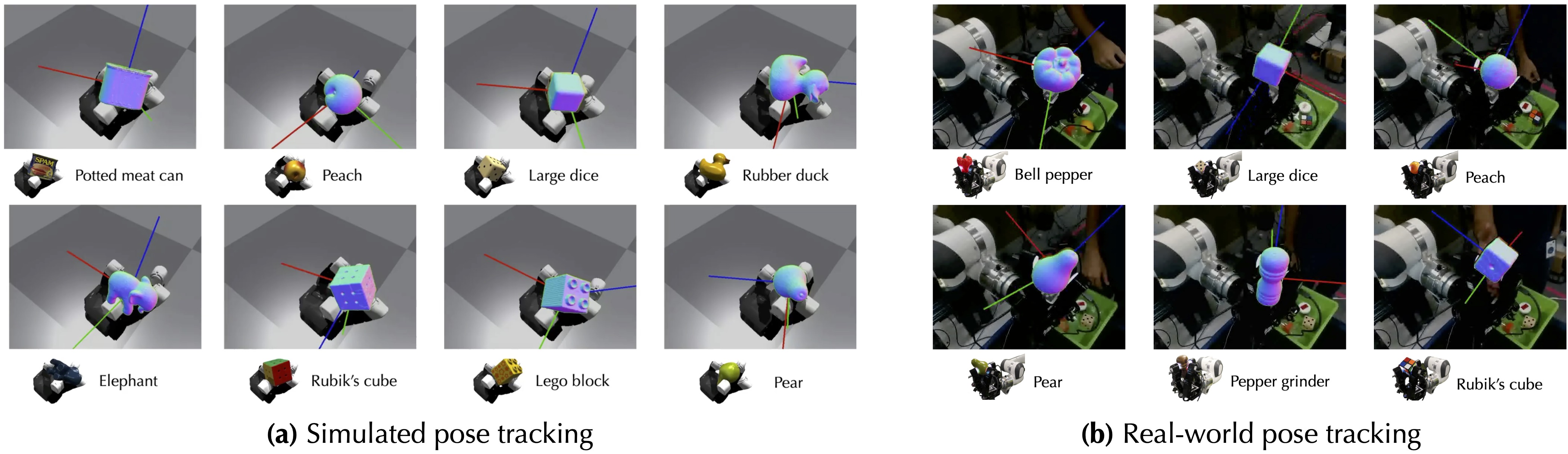}
    \caption{\textbf{Further visualizations of neural tracking experiments.} These qualitatively complement the results from \Sec \ref{ssec:neural_pose} for both \textbf{(a)} simulated and \textbf{(b)} real-world experiments. }
    \label{fig:pose_tracking_suppl}
\end{figure}

\noindent\textbf{Additional neural tracking visualizations.} \Fig \ref{fig:pose_tracking_suppl} shows rendering results from the experiments in \Sec \ref{ssec:neural_pose} along with the pose axes. We see good alignment of the renderings when overlaid on the RGB camera frame. 

\begin{figure}[t]
    \thisfloatpagestyle{empty}
    \centering
    \includegraphics[width=0.9\columnwidth,keepaspectratio]{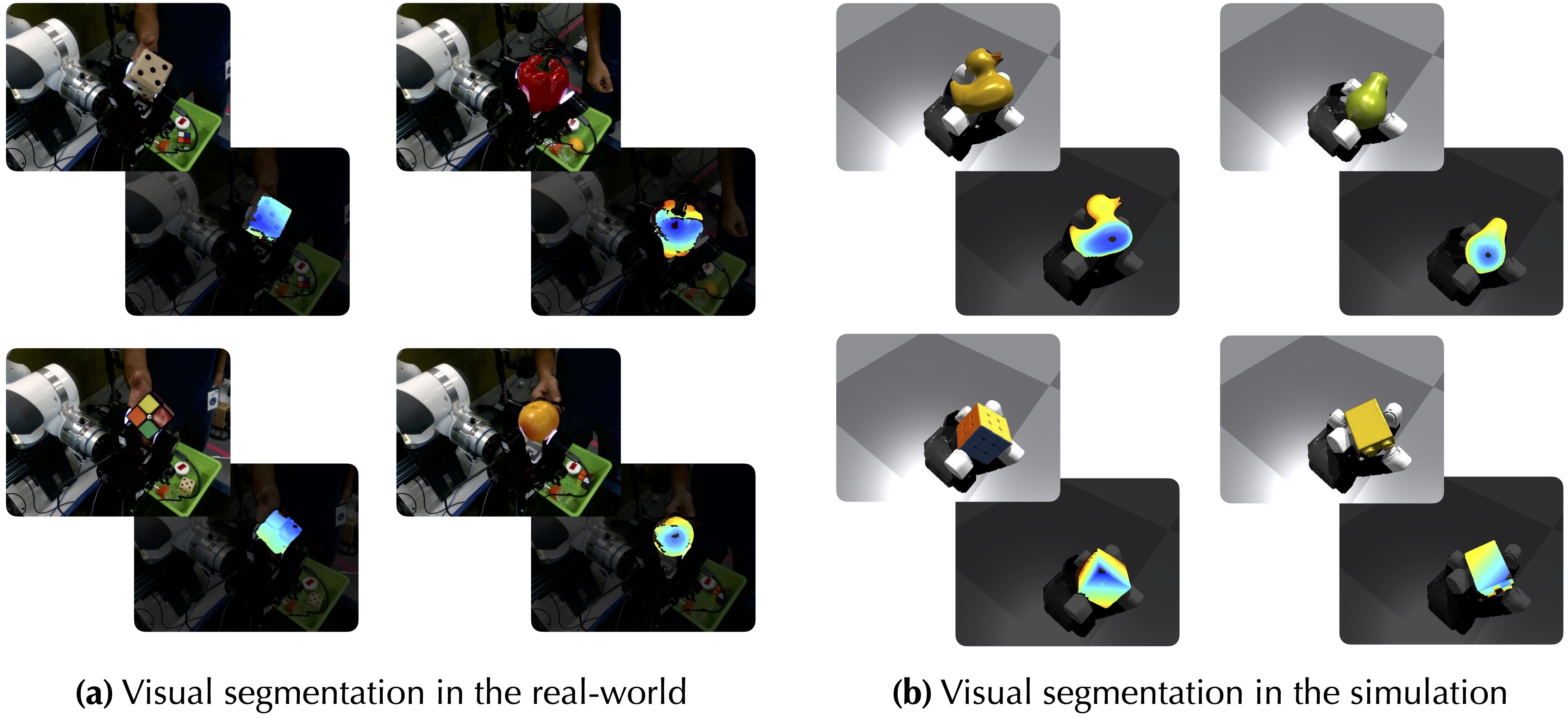}
    \caption{\textbf{Additional results on visual segmentation.} Our segmentation module can accurately singulate the in-hand object in both \textbf{(a)} real-world and \textbf{(b)} simulated image sequences.}
    \label{fig:sam_suppl}
\end{figure}

\noindent\textbf{Further visual segmentation results.} \Fig \ref{fig:sam_suppl} shows additional qualitative results of visual segmentation for \textbf{(a)} real-world and \textbf{(b)} simulated rotations sequences. 

\subsection{Illustrating the role of touch}
\label{ssec:role_touch}

\begin{figure}[t]
    \thisfloatpagestyle{empty}
    \centering
    \includegraphics[width=\columnwidth,keepaspectratio]{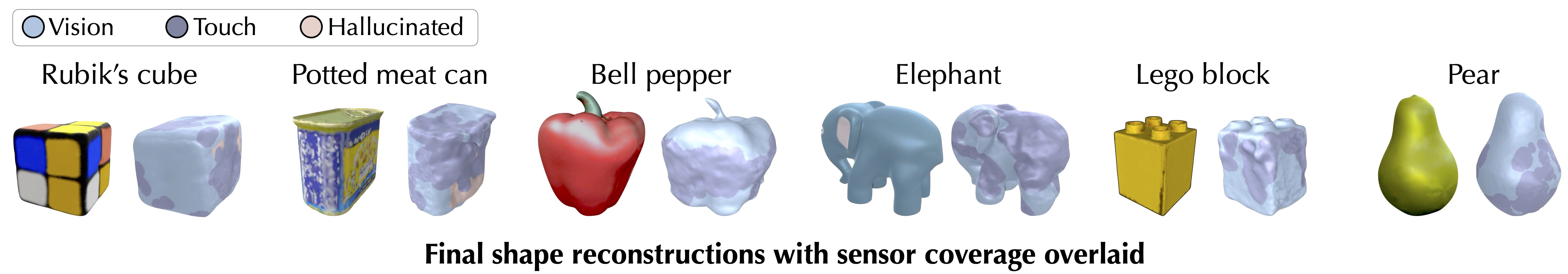}
    \caption{Sensor coverage illustrated in final mesh reconstructions of select objects---indicating vision, touch, and hallucinated regions.}
    \label{fig:sensor_meshes}
\end{figure}

\noindent\textbf{Sensor coverage visualized in SLAM.}  To illustrate the complementary nature of touch and vision, we color the reconstructed mesh regions based on their dominant sensing modality in \Fig \ref{fig:sensor_meshes}. After running the SLAM experiments in \Sec \ref{ssec:neural_slam}, we first run marching-cubes on the final neural SDF. In the resultant mesh, we assign each vertices color based on if vision or touch is the nearest pointcloud measurement to it. In the case where there is no vision or touch pointcloud within a $5\,\text{mm}$ radius, it is assigned as a \textit{hallucinated} vertex. This is a demonstrable advantage of neural SDFs, where the network can extrapolate well based on information in the neighborhood of the query point. From the meshes in \Fig \ref{fig:sensor_meshes} we see that while vision gets broad coverage of each object, there is considerable tactile signal from the interaction utilized for shape estimation. 

\begin{figure}[t]
    \thisfloatpagestyle{empty}
    \centering
    \includegraphics[width=\columnwidth,keepaspectratio]{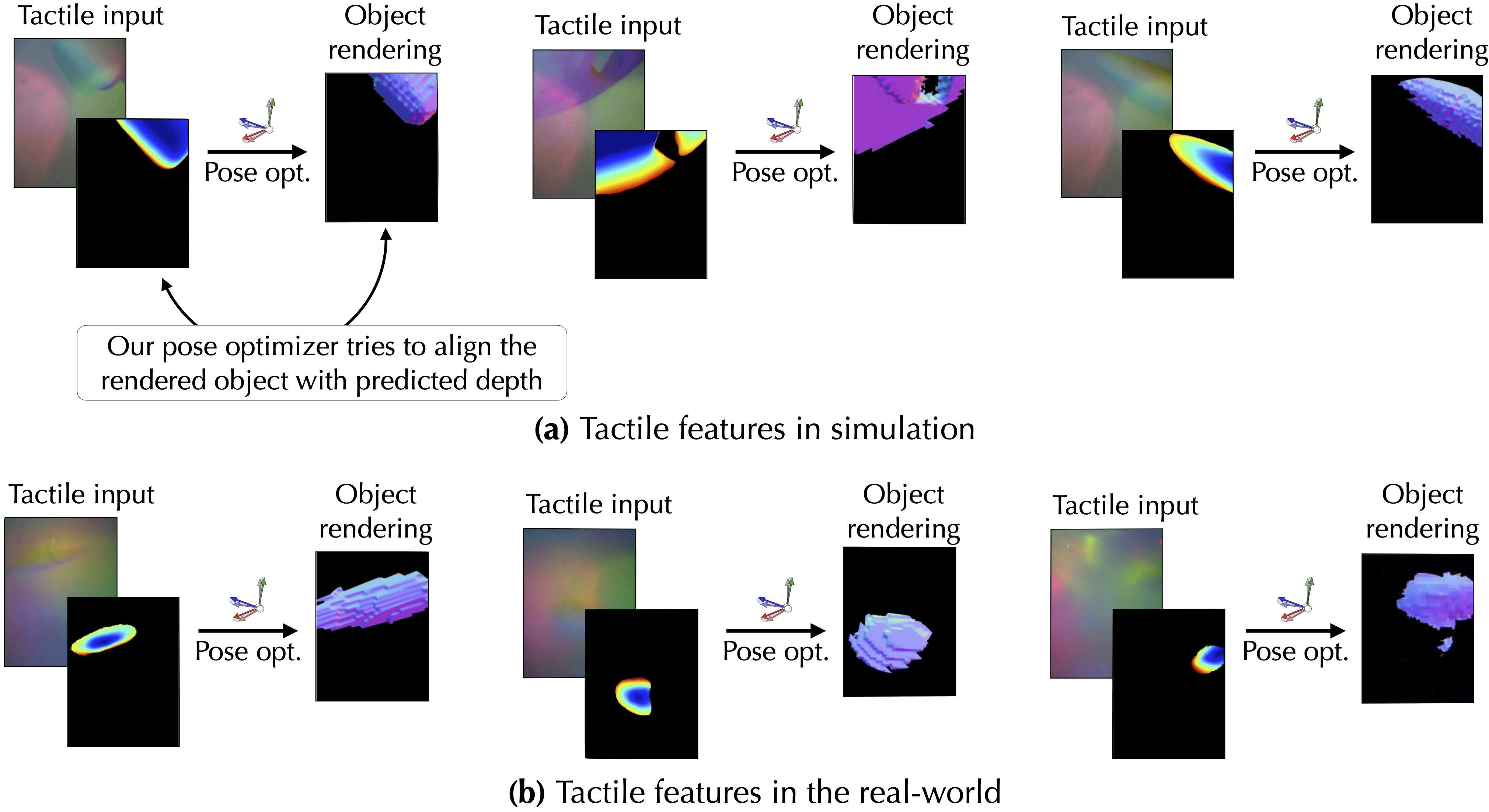}
    \caption{Six examples of tactile images compared against the neural field. We see that our tactile pose optimizer matches the predicted local geometry with the neural surface rendering. Thus, patches and edges predicted by touch appear in the rendering as well.}
    \label{fig:tactile_features}
\end{figure}

\noindent\textbf{Touch aligns local geometries with predicted depth.} As described in \Sec \ref{ssec:pose_optimizer}, the pose optimizer \textit{inverts} the neural field to back-propagate a loss in pose space~\cite{yen2021inerf, sucar2021imap, zhu2022nice}. This has been illustrated in work such as iNeRF~\cite{yen2021inerf}, where the rendered neural field attempts to match the image measurements via updates to the se(3) Lie algebra of the camera pose. As our framework leverages the idea that vision-based touch is just another perspective camera, we show how the rendered neural field matches with tactile depth features in \Fig \ref{fig:tactile_features}. 

Each RGB image is first passed through the tactile transformer (\Sec \ref{sssec:dpt_tactile}) to output a predicted tactile depthmap. Our pose optimizer aligns the neural rendering of the surface with the measured depthmap, based on 3D samples from the measured depthmap. Thus we can see that both in simulation (\Fig \ref{fig:tactile_features} (a)) and the real-world (\Fig \ref{fig:tactile_features} (b)), the edge and patch features predicted match well with the rendered object.